\pdfoutput=1
\def\year{2019}\relax
\documentclass[letterpaper]{article}
\usepackage{aaai19} 
\usepackage{times} 
\usepackage{helvet} 
\usepackage{courier} 
\usepackage{url}
\usepackage{graphicx}
\usepackage{amsmath}
\usepackage{graphicx}
\usepackage{subcaption}
\usepackage{tikz}
\usepackage{cuted}
\usetikzlibrary{bayesnet, calc, positioning}
\usepackage{booktabs}
\usepackage{wrapfig}
\usepackage{adjustbox}
\usepackage{enumitem}

\usepackage[utf8]{inputenc} 
\usepackage[T1]{fontenc}    
\usepackage{url}            
\usepackage{booktabs}       
\usepackage{amsfonts}       
\usepackage{nicefrac}       
\usepackage{stfloats}
\newcommand{\elboa}{\mathcal{L}^{(\mathrm{ind})}}
\newcommand{\elbob}{\mathcal{L}^{(\mathrm{MoE})}}
\newcommand{\elboc}{\mathcal{L}^{(\mathrm{PoE})}}
\newcommand{\elboac}{\mathcal{L}^{(\mathrm{hybrid})}}

\usepackage{todonotes}

\newcommand{\z}{\mathbf{z}}
\newcommand{\x}{\mathbf{x}}
\newcommand{\allx}{\mathbf{x}}

\newcommand{\alltheta}{\theta}
\newcommand{\allphi}{\phi}

\newcommand{\X}{\mathbf{X}}

\newcommand{\given}{\nonscript\,\vert\nonscript\,}
 \title{Multi-Source Neural Variational Inference}
\author{
  Richard Kurle \\
  Department of Informatics \\  %
  Technical University of Munich, \\
  Data:Lab, Volkswagen Group \\
  80805 Munich, Germany \\
  \texttt{richard.kurle@tum.de} \\
  \And
  Stephan G\"{u}nnemann \\
  Department of Informatics \\  %
  Technical University of Munich \\
  \texttt{guennemann@in.tum.de} \\
  \And
  Patrick van der Smagt \\
  Data:Lab, Volkswagen Group \\
  80805 Munich, Germany 
}
\begin{document}
\maketitle
\begin{abstract}
Learning from multiple sources of information is an important problem in machine-learning research. 
The key challenges are learning representations and formulating inference methods that take into account the complementarity and redundancy of various information sources.
In this paper we formulate a variational autoencoder based multi-source learning framework in which each encoder is conditioned on a different information source. 
This allows us to relate the sources via the shared latent variables by computing divergence measures between individual source's posterior approximations. 
We explore a variety of options to learn these encoders and to integrate the beliefs they compute into a consistent posterior approximation. 
We visualise learned beliefs on a toy dataset and evaluate our methods for learning shared representations and structured output prediction, showing trade-offs of learning separate encoders for each information source. 
Furthermore, we demonstrate how conflict detection and redundancy can increase robustness of inference in a multi-source setting. 
\end{abstract}
\section{Introduction}\label{sec:introduction}
An essential feature of most living organisms is the ability to process, relate, and integrate information coming from a vast number of sensors and eventually from memories and predictions \cite{Stein1993}. 
While integrating information from complementary sources enables a coherent and unified description of the environment, redundant sources are beneficial for reducing uncertainty and ambiguity. 
Furthermore, when sources provide conflicting information, it can be inferred that some sources must be unreliable. 

Replicating this feature is an important goal of multimodal machine learning \cite{baltruvsaitis2017multimodal}.
Learning joint representations of multiple modalities has been attempted using various methods, including neural networks \cite{NgiamKKNLN11}, probabilistic graphical models \cite{srivastava14b}, and canonical correlation analysis \cite{Andrew2013}. 
These methods focus on learning joint representations and multimodal sensor fusion. 
However, it is challenging to relate information extracted from different modalities. 
In this work, we aim at learning probabilistic representations that can be related to each other by statistical divergence measures as well as translated from one modality to another. 
We make no assumptions about the nature of the data (i.e. multimodal or multi-view) and therefore adopt a more general problem formulation, namely learning from multiple \emph{information sources}. 

Probabilistic graphical models are a common choice to address the difficulties of learning from multiple sources by modelling relationships between information sources---i.e., observed random variables---via unobserved, random variables. 
Inferring the hidden variables is usually only tractable for simple linear models.
For nonlinear models, one has to resort to approximate Bayesian methods. 
The variational autoencoder (VAE) \cite{KingmaW13,RezendeMW14} is one such method, combining neural networks and variational inference for latent-variable models (LVM). 

We build on the VAE framework, jointly learning the generative and inference models from multiple information sources.
In contrast to the VAE, we encapsulate individual inference models into separate ``modules''.
As a result, we obtain multiple posterior approximations, each informed by a different source. 
These posteriors represent the belief over the \emph{same} latent variables of the LVM, conditioned on the available information in the respective source. 

Modelling beliefs individually---but coupled by the generative model---enables computing meaningful quantities such as measures of surprise, redundancy, or conflict between beliefs. 
Exploiting these measures can in turn increase the robustness of the inference models.
Furthermore, we explore different methods to integrate arbitrary subsets of these beliefs, to approximate the posterior for the respective subset of observations. 
We essentially modularise neural variational inference in the sense that information sources and their associated encoders can be flexibly interchanged and combined after training.
\section{Background---Neural variational inference} \label{sec:background}
Consider a dataset $\X = \{ \x^{(n)} \}^{N}_{n=1} $ of $N$ i.i.d.~samples of some random variable $\x$ and the following generative model:
\begin{equation*} 
p_{\theta}(\x^{(n)}) 
= \int p_{\theta}(\x^{(n)} \given \z^{(n)} ) \, p(\z^{(n)}) \, d\z^{(n)},
\end{equation*}
where $\theta$ are the parameters of a neural network, defining the conditional distribution between latent and observable random variables $\z$ and $\x$ respectively.
The variational autoencoder \cite{KingmaW13,RezendeMW14} is an approximate inference method that enables learning the parameters of this model by optimising an evidence lower bound (ELBO) to the log marginal likelihood. 
A second neural network with parameters $\phi$ defines the parameters of an approximation $q_{\phi}(\z \given \x)$ of the posterior distribution. 
Since the computational cost of inference for each data point is shared by using a recognition model, some authors refer to this form of inference as amortised or neural variational inference \cite{GershmanG14,MnihG14}. 

The importance weighted autoencoder \cite{BurdaGS15} (IWAE) generalises the VAE by using a multi-sample importance weighting estimate of the log-likelihood. 
The IWAE ELBO is given as:
\begin{equation*}
\ln p_{\theta}(\mathbf{x}^{(n)}) 
\geq \mathbb{E}_{\z^{(n)}_{1:K} \sim q_{\phi}(\z^{(n)} \given \x^{(n)})} \Big{[} 
\ln \frac{1}{K} \sum_{k=1}^{K} w^{(n)}_{k} \Big{]} ,
\end{equation*}
where $K$ is the number of importance samples, and $w^{(n)}_{k}$ are the importance weights:
\begin{equation*}
w^{(n)}_{k} = \frac{p_{\theta}(\x^{(n)} \given \z^{(n)}_{k}) \, p(\z^{(n)}_{k})}{q_{\phi}(\z^{(n)}_{k} \given \x^{(n)})}.
\end{equation*}
Besides achieving a tighter lower bound, the IWAE was motivated by noticing that a multi-sample estimate does not require all samples from the variational distribution to have a high posterior probability. 
This enables the training of a generative model using samples from a variational distribution with higher uncertainty.
Importantly, this distribution need not be the posterior of all observations in the generative model. 
It can be a good enough proposal distribution, i.e.~the belief from a partially-informed source. 
\section{Multi-source neural variational inference}\label{sec:MIWVI}
We are interested in datasets consisting of tuples $\{\x^{(n)} = ( \x_{1}^{(n)}, \dots,~\x_{M}^{(n)}) \}^{N}_{n=1}$, we use $m \in \{ 1,\dots,M \}$ to denote the index of the source. 
Each observation $\x_{m}^{(n)} \in \mathbb{R}^{D_{m}}$ may be embedded in a different space but is assumed to be generated from the same latent state $\z^{(n)}$. 
Therefore, each $\x_{m}^{(n)}$ corresponds to a different, potentially limited source of information about the underlying state $\z^{(n)}$. 
From now on we will refer to $\x_{m}$ in the generative model as observations and the same $\x_{m}$ in the inference model as information sources. 

We model each observation $\x_{m}$ in the generative model with a distinct set of parameters $\theta_{m}$, although some parameters could be shared. 
The likelihood function is given as: 
\begin{equation*}
p_{\alltheta}(\x^{(n)} \given \z^{(n)}) = \prod_{m=1}^{M} p_{\theta_{m}}\big(\x^{(n)}_{m} \given \z^{(n)}\big).
\end{equation*}
For inference, the VAE conditions on all observable data $\x^{(n)}$. 
However, one can condition (amortize) the approximate posterior distribution on any set of information sources.
In this paper we limit ourselves to $\x^{(n)}_S, \: S \subset \{1, \ldots, M\}$. 
An approximate posterior distribution $q_{\phi_{S}}(\z^{(n)} \given \x_{S}^{(n)})$ may then be interpreted as the belief of the respective information sources about the latent variables, underlying the generative process. 

In contrast to the VAE, we want to calculate the beliefs from \emph{different} information sources individually, compare them, and eventually integrate them. 
In the following, we address each of these desiderata.

%
\begin{figure*}
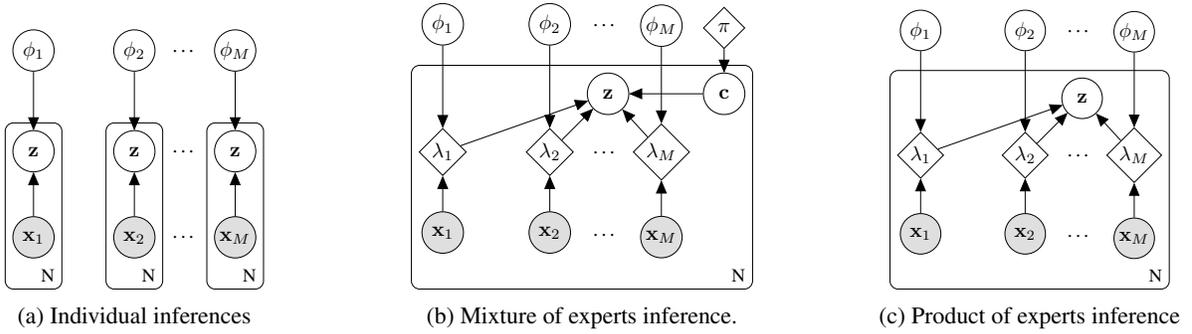

  \centering
  \begin{subfigure}[t]{0.31\textwidth}
  \centering
  \resizebox{0.65\linewidth}{!}{
   \tikz{ %
    \node[latent] (Z1) {$\z$} ; %
    \node[latent, right=of Z1] (Z2) {$\z$} ; %
    \node[latent, right=of Z2] (ZM) {$\z$} ; %
    \node at ($(Z2)!.5!(ZM)$) {\ldots};
    
    \node[obs, below=0.75 of Z1 ] (X1) {$\x_{1}$} ; %
    \node[obs, below=0.75 of Z2] (X2) {$\x_{2}$} ; %
    \node[obs, below=0.75 of ZM] (XM) {$\x_{M}$} ; 
    \node at ($(X2)!.5!(XM)$) {\ldots};    
    
    \plate{plate_locals_1}{(X1)(Z1)}{N} ;
    \plate{plate_locals_2}{(X2)(Z2)}{N} ;
    \plate{plate_locals_M}{(XM)(ZM)}{N} ;
    
    \node[latent, above=of Z1] (phi1) {$\phi_{1}$} ;
    \node[latent, above=of Z2] (phi2) {$\phi_{2}$} ;
    \node[latent, above=of ZM] (phiM) {$\phi_{M}$} ;
    \node at ($(phi2)!.5!(phiM)$) {\ldots};    
    
    \edge {phi1} {Z1} ;
    \edge {phi2} {Z2} ;      
    \edge {phiM} {ZM} ;      
    
    \edge {X1} {Z1} ; %
    \edge {X2} {Z2} ; %
    \edge {XM} {ZM} ; %
    }
  }
  \caption{Individual inferences}
  \label{fig:inference_individual}
  \end{subfigure}
  \quad
  \begin{subfigure}[t]{0.31\textwidth}
  \centering
  \resizebox{0.85\linewidth}{!}{
  \tikz{ %
    \node[det] (lambda1) {$\lambda_{1}$} ; %
    \node[det, right=of lambda1] (lambda2) {$\lambda_{2}$} ; %
    \node[det, right=of lambda2] (lambdaM) {$\lambda_{M}$} ; %
    \node at ($(lambda2)!.5!(lambdaM)$) {\ldots};
	
	\node[latent, above right=0.75 of lambda2] (Z) {$\z$} ;
	\node[latent, right=1.25 of Z] (C) {$\mathbf{c}$} ;
	\node[det, above=0.4 of C] (pi) {$\mathbf{\pi}$} ;
    
    \node[obs, below=0.6 of lambda1] (X1) {$\x_{1}$} ; %
    \node[obs, below=0.6 of lambda2] (X2) {$\x_{2}$} ; %
    \node[obs, below=0.6 of lambdaM] (XM) {$\x_{M}$} ; 
    \node at ($(X2)!.5!(XM)$) {\ldots};    
    
    \plate{plate}{(X1)(X2)(XM)(lambda1)(lambda2)(lambdaM)(Z)(C)}{N} ;
    
    \node[latent, above=1.40 of lambda1] (phi1) {$\phi_{1}$} ;
    \node[latent, above=1.40 of lambda2] (phi2) {$\phi_{2}$} ;
    \node[latent, above=1.30 of lambdaM] (phiM) {$\phi_{M}$} ;
    \node at ($(phi2)!.5!(phiM)$) {\ldots};
    
    \edge {phi1} {lambda1} ;
    \edge {phi2} {lambda2} ;      
    \edge {phiM} {lambdaM} ;      
    
    \edge {X1} {lambda1} ; %
    \edge {X2} {lambda2} ; %
    \edge {XM} {lambdaM} ; %
    
    \edge {lambda1} {Z}; %
    \edge {lambda2} {Z}; %
    \edge {lambdaM} {Z}; %
    \edge {C} {Z}
    \edge {pi}{C}
    }
  }
  \caption{Mixture of experts inference. }
  \label{fig:inference_moe}
  \end{subfigure}
  \quad
  \begin{subfigure}[t]{0.31\textwidth}
  \centering
  \resizebox{0.70\linewidth}{!}{
  \tikz{ %
    \node[det] (lambda1) {$\lambda_{1}$} ; %
    \node[det, right=of lambda1] (lambda2) {$\lambda_{2}$} ; %
    \node[det, right=of lambda2] (lambdaM) {$\lambda_{M}$} ; %
    \node at ($(lambda2)!.5!(lambdaM)$) {\ldots};
    	
	\node[latent, above right=0.75 of lambda2] (Z) {$\z$} ;
    
    \node[obs, below=0.60 of lambda1 ] (X1) {$\x_{1}$} ; 
    \node[obs, below=0.60 of lambda2] (X2) {$\x_{2}$} ; 
    \node[obs, below=0.60 of lambdaM] (XM) {$\x_{M}$} ; 
    \node at ($(X2)!.5!(XM)$) {\ldots};    
    
    \plate{plate}{(X1)(X2)(XM)(lambda1)(lambda2)(lambdaM)(Z)}{N} ;
    
    \node[latent, above=1.40 of lambda1] (phi1) {$\phi_{1}$} ;
    \node[latent, above=1.40 of lambda2] (phi2) {$\phi_{2}$} ;
    \node[latent, above=1.30 of lambdaM] (phiM) {$\phi_{M}$} ;
    \node at ($(phi2)!.5!(phiM)$) {\ldots};    
    
    \edge {phi1} {lambda1} ;
    \edge {phi2} {lambda2} ;      
    \edge {phiM} {lambdaM} ;      
    
    \edge {X1} {lambda1} ; %
    \edge {X2} {lambda2} ; %
    \edge {XM} {lambdaM} ; %
    
    \edge {lambda1} {Z}; %
    \edge {lambda2} {Z}; %
    \edge {lambdaM} {Z}; %
    }
  }
  \caption{Product of experts inference}
  \label{fig:inference_poe}
  \end{subfigure}
\caption{Graphical models of inference models. White circles denote hidden random variables, grey-shaded circles---observed random variables, diamonds---deterministic variables. N is the number of i.i.d.~samples in the dataset. 
To better distinguish the mixture or product of expert models from an IWAE with hard-wired integration in a neural-network layer, we explicitly draw the deterministic variables $\lambda_{1}, \dots , \lambda_{M}$, denoting the parameters of the variational distributions.}
\end{figure*}

\subsection{Learning individual beliefs}\label{sec:learning_individual_beliefs}
In order to learn individual inference models as in Fig.~\ref{fig:inference_individual}, we propose an average of $M$ ELBOs, one for each information source and its respective inference model. 
The resulting objective is an ELBO to the log marginal likelihood itself and referred to as $\elboa$:
\begin{equation} \label{eq:boundA}
\elboa =:
\sum_{m=1}^{M} \pi_{m} \mathbb{E}_{\z^{(n)}_{1:K} \sim q_{\phi_m}
\big(\z^{(n)} \given \x_{m}^{(n)}\big)} 
\Big{[} 
\ln \frac{1}{K} \sum_{k=1}^{K} w^{(n)}_{m, k} \Big{]}, 
\end{equation}
with
\begin{equation*}
w^{(n)}_{m, k} = \frac{p_{\alltheta}\big(\allx^{(n)} \given \z^{(n)}_{k}\big) p\big( \z^{(n)}_{k} \big) }{q_{\phi_m}\big(\z_{k}^{(n)} \given \x_{m}^{(n)}\big) }.
\end{equation*}
The indices $n$, $m$ and $k$ refer to the data sample, information source, and importance sample index. 
The factors $\pi_{m}$ are the weights of the ELBOs, satisfying $0 \leq \pi_{m} \leq 1$ and $\sum_{m=1}^{M}\pi_{m} = 1$. 
Although the $\pi_{m}$ could be inferred, we set $\pi_{m} = 1/M,~\forall m$. 
This ensures that all parameters $\phi_{m}$ are optimised individually to their best possible extent instead of down-weighting less informative sources. 

Since we are dealing with partially-informed encoders $q_{\phi_{m}}(\z^{(n)} \given \x_{m}^{(n)})$ instead of $q_{\allphi}(\z^{(n)} \given \allx^{(n)})$, the beliefs can be more uncertain than the posterior of all observations $\allx$. 
This in turn degrades the generative model, as it requires samples from the posterior distribution. 
We found that the generative model becomes biased towards generating averaged samples rather than samples from a diverse, multimodal distribution. 
This issue arises in VAE-based objectives, irrespective of the complexity of the variational family, because each Monte-Carlo sample of latent variables must predict all observations. 
To account for this, we propose to use importance sampling estimates of the log-likelihood (see Sec.~\ref{sec:background}). 
The importance weighting and sampling-importance-resampling can be seen as feedback from the observations, allowing to approximate the true posterior even with poorly informed beliefs. 

\subsection{Comparing beliefs}
Encapsulating individual inferences has an appealing advantage compared to an uninterpretable, deterministic combination within a neural network: 
Having obtained multiple beliefs w.r.t.\ the same latent variables, each informed by a distinct source, we can calculate meaningful quantities to relate the sources. 
Examples are measures of redundancy, surprise, or conflict. 
Here we focus on the latter. 

Detecting conflict between beliefs is crucial to avoid false inferences and thus increase robustness of the model. 
Conflicting beliefs may stem from conflicting data or from unreliable (inference) models. 
The former is a form of data anomaly, e.g.~due to a failing sensor. 
An unreliable model on the other hand may result from model misspecification or optimisation problems, i.e.\ due to the approximation or amortisation gap, respectively \cite{CremerC2018a}. 
Distinguishing between the two causes of conflict is challenging however and requires evaluating the observed data under the likelihood functions.

Previous work has used the ratio of two KL divergences as a criterion to detect a conflict between a subjective prior and the data \cite{Bousquet2008}.
The nominator is the KL between the posterior and the subjective prior, and denominator is the KL between posterior and a non-informative reference prior. 
The two KL divergences measure the information gain of the posterior---induced by the evidence---w.r.t.~the subjective prior and the non-informative prior, respectively. 
The decision criterion for conflict is a ratio greater than 1. 

We propose a similar ratio, replacing the subjective prior with $q_{\phi_{m}}$ and taking the prior as reference:
\begin{equation}\label{eq:conflict}
\mathrm{c}(m~||~m') = \frac
{D_{\mathrm{KL}}\big(q_{\phi_{m'}}(\z \given \x_{m'}) ~||~ q_{\phi_{m}}(\z \given \x_{m})\big)}
{D_{\mathrm{KL}}\big(q_{\phi_{m'}}(\z \given \x_{m'}) ~||~ p(\z)\big)}.
\end{equation}
This measure has the property that it yields high values if the belief of source $m$ is significantly more certain than that of $m'$. 
This is desirable for sources with redundant information. 
For complementary information sources other conflict measures, e.g.\ the measure defined in \cite{Dahl2007}, may be more appropriate. 

\subsection{Integrating beliefs} \label{sec:Information_Integration}
So far, we have shown how to learn separate beliefs from different sources and how to relate them. 
However, we have not readily integrated the information from these sources. 
This can be seen by noticing that the gap between $\elboa$ and the log marginal likelihood is significantly larger compared to an IWAE with an unflexible, hard-wired combination (see supplementary material). 
Here we propose two methods to integrate the beliefs $q_{\phi_{m}}(\z \given x_{m})$ to an integrated belief $q_{\allphi}(\z \given \allx)$.

\subsubsection{Disjunctive integration---Mixture of Experts}\label{sec:MoE}
One approach to combine individual beliefs is by treating them as alternatives, which is justified if some (but not all) sources or their respective models are unreliable or in conflict \cite{Khaleghi2013}. 
We propose a mixture of experts (MoE) distribution, where each component is the belief, informed by a different source. 
The corresponding graphical model for inference is shown in Fig.~\ref{fig:inference_moe}. 
As in Sec.~\ref{sec:learning_individual_beliefs}, the variational parameters are each predicted from one source individually without communication between them. 
The difference is that each $q_{\phi_{m}}(\z \given \x_{m})$ is considered as a mixture component, such that the whole mixture distribution approximates the true posterior.

Instead of learning individual beliefs $q_{\phi_{m}}(\z \given \x_{m})$ by optimising $\elboa$ and integrating them subsequently into a combined $q_{\allphi}(\z \given \allx)$, we can design an objective function for learning the MoE posterior directly.
We refer to the corresponding ELBO as $\elbob$.
It differs from $\elboa$ only by the denominator of the importance weights, using the mixture distribution with component weights $\pi_{m}$:
\begin{equation*}
w^{(n)}_{m, k} = \frac{p_{\alltheta} \big( \allx^{(n)} \given \z^{(n)}_{k}\big) \, p \big( \z^{(n)}_{k} \big) }{\sum_{m'=1}^{M} \pi_{m'} q_{\phi_{m'}} \big( \z_{k}^{(n)} \given \x_{m'}^{(n)} \big) },
\end{equation*}

\subsubsection{Conjunctive integration---Product of Experts}\label{sec:PoE}
Another option for combining beliefs are conjunctive methods, treating each belief as a constraint.
These are applicable in the case of equally reliable and independent evidences \cite{Khaleghi2013}.
This can be seen by inspecting the mathematical form of the posterior distribution of all observations. 
Applying Bayes' rule twice reveals that the true posterior of a graphical model with conditionally independent observations can be decomposed as a product of experts \cite{HintonG2002} (PoE):
\begin{equation} \label{eq:posterior_product}
p(\z \given  \allx) 
= \frac{\prod_{m'=1}^{M} p(\x_{m'})}{p(\allx)} \cdot
p(\z) \cdot
\prod_{m=1}^{M} \frac{p(\z \given \x_{m})}{p(\z)} .
\end{equation}
We propose to approximate Eq.~(\ref{eq:posterior_product}) by replacing the true posteriors of single observations $p(\z \given \x_{m})$ by the variational distributions $q_{\phi_{m}}(\z \given \x_{m})$, obtaining the inference model shown in Fig.~\ref{fig:inference_poe}. 
In order to make the PoE distribution computable, we further assume that the variational distributions and the prior are conjugate distributions in the exponential family. 
Probability distributions in the exponential family have the well-known property that their product is also in the exponential family. 
Hence, we can calculate the normalisation constant in Eq.~(\ref{eq:posterior_product}) from the natural parameters. 
In this work, we focus on the popular case of normal distributions. 
For the derivation of the natural parameters and normalisation constant, we refer to the supplementary material.

Analogous to Sec.~\ref{sec:MoE}, we can design an objective to learn the PoE distribution directly, rather than integrating individual beliefs.
We refer to the corresponding ELBO as $\elboc$:
\begin{equation}
\elboc =: 
\mathbb{E}_{\z^{(n)}_{1:K} \sim q_{\allphi}
\big( \z^{(n)} \given \allx^{(n)} \big)} 
\Big{[} 
\ln \frac{1}{K} \sum_{k=1}^{K} w^{(n)}_{k} \Big{]}, 
\end{equation}
where $w^{(n)}_{k}$ are the standard importance weights as in the IWAE and where $q_{\allphi}
( \z^{(n)} \given \allx^{(n)})$ is the PoE inference distribution. 
However, the natural parameters of the individual normal distributions are not uniquely identifiable by the natural parameters of the integrated normal distribution. Thus, optimising $\elboc$ leads to inseparable individual beliefs. 
To account for this, we propose a hybrid between individual and integrated inference distribution:
\begin{equation}
\elboac = \lambda_{1} \elboa + \lambda_{2} \elboc, 
\end{equation}
where we choose $\lambda_{1} = \lambda_{2} = \frac{1}{2}$ in practice for simplicity. 

In Sec.~\ref{sec:experiments} we evaluate the proposed integration methods both as learning objectives, and for integrating the beliefs obtained by optimising $\elboa$ or $\elboac$. 
Note again however, that $\elboc$ or $\elboac$ assume conditionally independent observations and equally reliable sources. 
In contrast, $\elboa$ makes no assumptions about the structure of the generative model. 
This allows for any choice of appropriate integration method after learning. 
\section{Related Work}\label{sec:related_work}
%
Canonical correlation analysis (CCA) \cite{Hotelling1936} is an early attempt to examine the relationship between two sets of variables.
CCA and nonlinear variants \cite{Shon2005,Andrew2013,Feng2015DeepCR} propose projections of pairs of features such that the transformed representations are maximally correlated. 
CCA variants have been widely used for learning from multiple information sources \cite{Hardoon2004,Rasiwasia2010}. 
These methods have in common with ours, that they learn a common representational space for multimodal data. 
Furthermore, a connection between linear CCA and probabilistic graphical models has been shown \cite{Bach2005}. 

Dempster-Shafer theory \cite{dempster1967,shafer1976} is a widely used framework for integration of uncertain information. 
Similar to our PoE integration method, Dempster's rule of combination takes the pointwise product of belief functions and normalises subsequently. 
Due to apparently counterintuitive  results  obtained when dealing with conflicting information \cite{Zadeh1986}, the research community proposed various measures to detect conflicting belief functions and proposed alternative integration methods. 
These include disjunctive integration methods \cite{s16091509,DENOEUX2008234,DengY2015,MurphyC2000}, similar to our MoE integration method. 

A closely related line of research is that of multimodal autoencoders \cite{NgiamKKNLN11} and multimodal Deep Boltzmann machines (DBM) \cite{srivastava14b}. 
Multimodal autoencoders use a shared representation for input and reconstructions of different modalities. 
Since multimodal autoencoders learn only deterministic functions, the interpretability of the representations is limited. 
Multimodal DBMs on the other hand learn multimodal generative models with a joint representation between the modalities. 
However, DBMs have only been shown to work on binary latent variables and are notoriously hard to train. 

More recently, variational autoencoders were applied to multimodal learning \cite{Suzuki2016}. 
Their objective function maximises the ELBO using an encoder with hard-wired sources and additional KL divergence loss terms to train individual encoders. 
The difference to our methods is that we maximise an ELBO for which we require only $M$ individual encoders. 
We may then integrate the beliefs of arbitrary subsets of information sources after training. 
In contrast, the method in \cite{Suzuki2016} would require a separate encoder for each possible combination of sources. 
Similarly, \cite{Vedantam2017} first trains a generative model with multiple observations, using a fully-informed encoder. 
In a second training stage, they freeze the generative model parameters and proceed by optimising the parameters of inference models which are informed by a single source. 
Since the topology of the latent space is fixed in the second stage, finding good weights for the inferenc models may be complicated.
 
Concurrently to this work, \cite{WuM2018} proposed a method for weakly-supervised learning from multimodal data, which is very similar to our hybrid method discussed in Sec.~\ref{sec:PoE}. 
Their method is based on the VAE, whereas we find it crucial to optimise the importance-sampling based ELBO to prevent the generative models from generating averaged conditional samples (see Sec.~\ref{sec:learning_individual_beliefs}). 
\section{Experiments}\label{sec:experiments}
We visualise learned beliefs on a 2D toy problem, evaluate our methods for structured prediction and demonstrate how our framework can increase robustness of inference. 
Model and algorithm hyperparameters are summarised in the supplementary material. 
\subsection{Learning beliefs from complementary information sources}
\begin{figure}[h!] \centering
\caption{Approximate posterior distributions and samples from the predicted likelihood function with and without integration of beliefs}
\begin{subfigure}{1.0\columnwidth} \centering
\includegraphics[width=1.0\columnwidth]{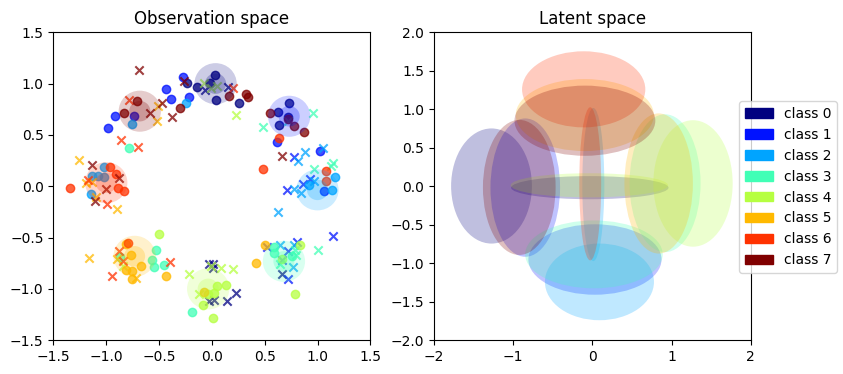}
\caption{Individual beliefs and their predictions.
\textbf{Left:} 8 coloured circles are centred at the 8 test inputs from a mixture of Gaussians toy dataset. 
The radii indicate 1 and 2 standard deviations of the normal distributions. 
The two types of markers represent generated data from random samples of one of the information sources (data axis 0 or 1).
\textbf{Right:} Corresponding individual beliefs. Ellipses show 1 standard deviation of the individual approximate posterior distributions.
}
\label{fig:POGOR_lat_and_pred_individual}
\end{subfigure}

\begin{subfigure}{1.0\columnwidth} \centering
\includegraphics[width=1.0\columnwidth]{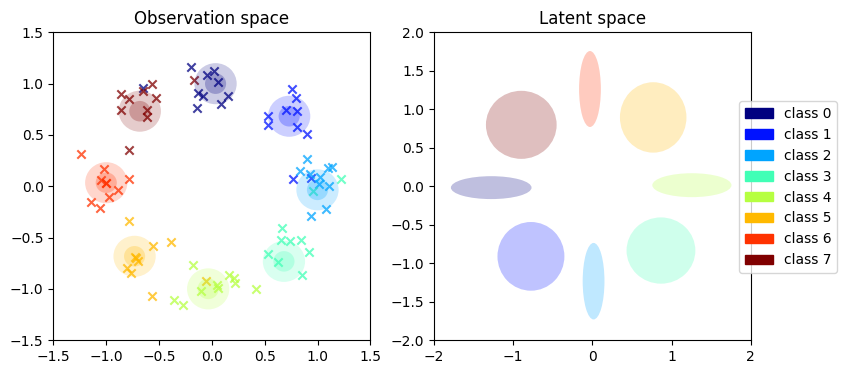}
\caption{Integrated belief and its predictions.}
\label{fig:POGOR_lat_and_pred_integrated}
\end{subfigure}
\end{figure}

We begin our experiments with a toy dataset with complementary sources. 
As a generative process, we consider a mixture of bi-variate normal distributions with 8 mixture components. 
The means of each mixture component are located on the unit circle with equidistant angles, and the standard deviations are $0.1$. 
To simulate complementary sources, we allow each source to perceive only one dimension of the data. 
As with all our experiments, we assume a zero-centred normal prior with unit variance and $\z \in \mathbb{R}^{2}$. 
We optimise $\elboa$ with two inference models $q_{\phi_{1}}(\z \given \x_{1})$, $q_{\phi_{2}}(\z \given  \x_{2})$, and two separate likelihood functions $p_{\theta_{1}}(\x_{1} \given \z)$, $p_{\theta_{2}}(\x_{2} \given \z)$. 
Fig.~\ref{fig:POGOR_lat_and_pred_individual}~(right) shows the beliefs of both information sources for 8 test data points. 
These test points are the means of the 8 mixture components of the observable data, rotated by $2^{\circ}$. 
The small rotation is only for visualisation purposes, since each source is allowed to perceive only one axis and would therefore produce indistinguishable beliefs for data points with identical values on the perceived axis. 
We visualise the two beliefs corresponding to the same data point with identical colours. 
The height and width of the ellipses correspond to the standard deviations of the beliefs. 
Fig.~\ref{fig:POGOR_lat_and_pred_individual}~(left) shows random samples in the observation space, generated from 10 random latent samples $\z \sim q_{\phi_{m}}(\z \given \x_{m})$ for each belief. 
The generated samples are colour-coded in correspondence to the figure on the right. 
The 8 circles in the background visualise the true data distribution with 1 and 2 standard deviations. 
The two types of markers distinguish the information sources $\x_{1}$ and $\x_{2}$ used for inference. 
As can be seen, the beliefs reflect the ambiguity as a result of perceiving a single dimension $x_{m}$. \footnote{The true posterior (of a single source) has two modes for most data points. The uni-modal (Gaussian) proposal distribution learns to cover both modes.}

Next we integrate the two beliefs using Eq.~(\ref{eq:posterior_product}).
The resulting integrated belief and generated data from random latent samples of the belief are shown in Figs.~\ref{fig:POGOR_lat_and_pred_integrated}~(right) and \ref{fig:POGOR_lat_and_pred_integrated}~(left) respectively. 
We can see that the integration resolves the ambiguity. 
In the supplementary material, we plot samples from the individual and integrated beliefs, before and after a sampling importance re-sampling procedure.

\subsection{Learning and inference of shared representations for structured prediction}\label{sec:experiments:representations_structured_prediction} 
Models trained with $\elboa$ or $\elboac$ can be used to predict structured data of any modality, conditioned on any available information source. 
Equivalently, we may impute missing data if modelled explicitly as an information source:
\begin{equation} \label{eq:conditional_prediction}
p(\x_{m} \given \x_{m'}) = \mathbb{E}_{\mathbf{z}\sim q_{\phi_{m'}} \big( \z \given \x_{m'} \big)} 
\Big[ p_{\theta_{m}}(\x_{m} \given \z) \Big].
\end{equation}
\subsubsection{MNIST variants}
We created 3 variants of MNIST \cite{lecun1998}, where we simulate multiple information sources as follows:
\begin{itemize} 
\item MNIST-TB: $\x_{1}$ perceives the \textbf{top} half and $\x_{2}$ perceives the \textbf{bottom} half of the image.
\item MNIST-QU: 4 information sources that each perceive \textbf{quarters} of the image. 
\item MNIST-NO: 4 information sources with independent bit-flip \textbf{noise} with $p = 0.05$. We use these 4 sources to amortise inference. In the generative model, we use the standard, noise-free digits as observable variables.
\end{itemize}
First, we assess how well individual beliefs can be integrated after learning, and whether beliefs can be used individually when learning them as integrated inference distributions. 
On all MNIST variants, we train 5 different models by optimising the objectives $\elboa$, $\elbob$, $\elboc$, and $\elboac$ with $K=16$, as well as $\elboac$ with $K=1$.  
All other hyperparameters are identical. 
We then evaluate each model under the 3 objectives $\elboa$, $\elbob$ and $\elboc$. 
For comparison, we also train a standard IWAE with hardwired sources on MNIST and on MNIST-NO with a single noisy source. 
The ELBOs on the test set are estimated using $K=16$ importance samples. 
The obtained estimates are summarised in Tab.~\ref{tab:mnist_elbo_table}. 
\begin{table}[h!]\centering
\caption{Negative evidence lower bounds on variants of randomly binarised MNIST. Lower is better.} 
\resizebox{1.0\columnwidth}{!} {
\begin{tabular}{lllllll}
&&& MNIST-TB &&& \\
\toprule
& $\elboa$ & $\elbob$ & $\elboc$ & $\elboac$ & $\elboac_{(K=1)}$ & IWAE \\
\midrule
$\elboa$ & \textbf{102.20} & 102.40 & 265.59 	& 104.03 & 108.97 & - \\
$\elbob$ & \textbf{101.51} & 101.82 & 264.48 	& 103.37	& 108.30 & - \\
$\elboc$ & 	94.38    & 94.39   & \textbf{87.59}  & 90.07 & 90.81  & 88.79 \\
\bottomrule
\end{tabular}
}
\resizebox{1.0\columnwidth}{!} {
\begin{tabular}{lllllll}
&&& MNIST-QU &&& \\
\toprule
& $\elboa$ & $\elbob$ & $\elboc$ & $\elboac$ & $\elboac_{(K=1)}$ & IWAE \\
\midrule
$\elboa$ & 120.46 & \textbf{120.37} & 447.67 & 129.63 & 140.61 & - \\
$\elbob$ & \textbf{119.10} & 119.98 & 446.02 	& 128.16	& 139.19 & - \\
$\elboc$ & 108.07    & 107.85   & \textbf{87.67}  & 89.20 & 90.17  & 88.79 \\
\bottomrule
\end{tabular}
}
\resizebox{1.0\columnwidth}{!} {
\begin{tabular}{lllllll}
&&& MNIST-NO &&& \\
\toprule
& $\elboa$ & $\elbob$ & $\elboc$ & $\elboac$ & $\elboac_{(K=1)}$ & IWAE \\
\midrule
$\elboa$ & \textbf{94.81} & 94.86 & 101.20 	& 96.27 & 95.31 & - \\
$\elbob$ & \textbf{93.98} & 94.03 & 100.36 	& 95.58	& 94.55 & - \\
$\elboc$ & 	94.52  & 94.65  & 92.27  & \textbf{92.21} & 94.49  & 94.95 \\
\bottomrule
\end{tabular}
}
\label{tab:mnist_elbo_table}
\end{table}
The results confirm that learning the PoE inference model directly leads to inseparable individual beliefs.
As expected, learning individual inference models and integrating them subsequently as a PoE comes with a tradeoff for $\elboc$, which is mostly due to the low entropy of the integrated distribution. 
On the other hand, optimising the model with $\elboac$ achieves good results for both individual and integrated beliefs.
On MNIST-NO, we can get an improvement of $2.74$ nats by integrating the beliefs of redundant sources, compared to the standard IWAE with a single source. 

\begin{figure}
	\centering
	\begin{subfigure}[t]{0.47\columnwidth} \centering
	\includegraphics[width=1.0\columnwidth]{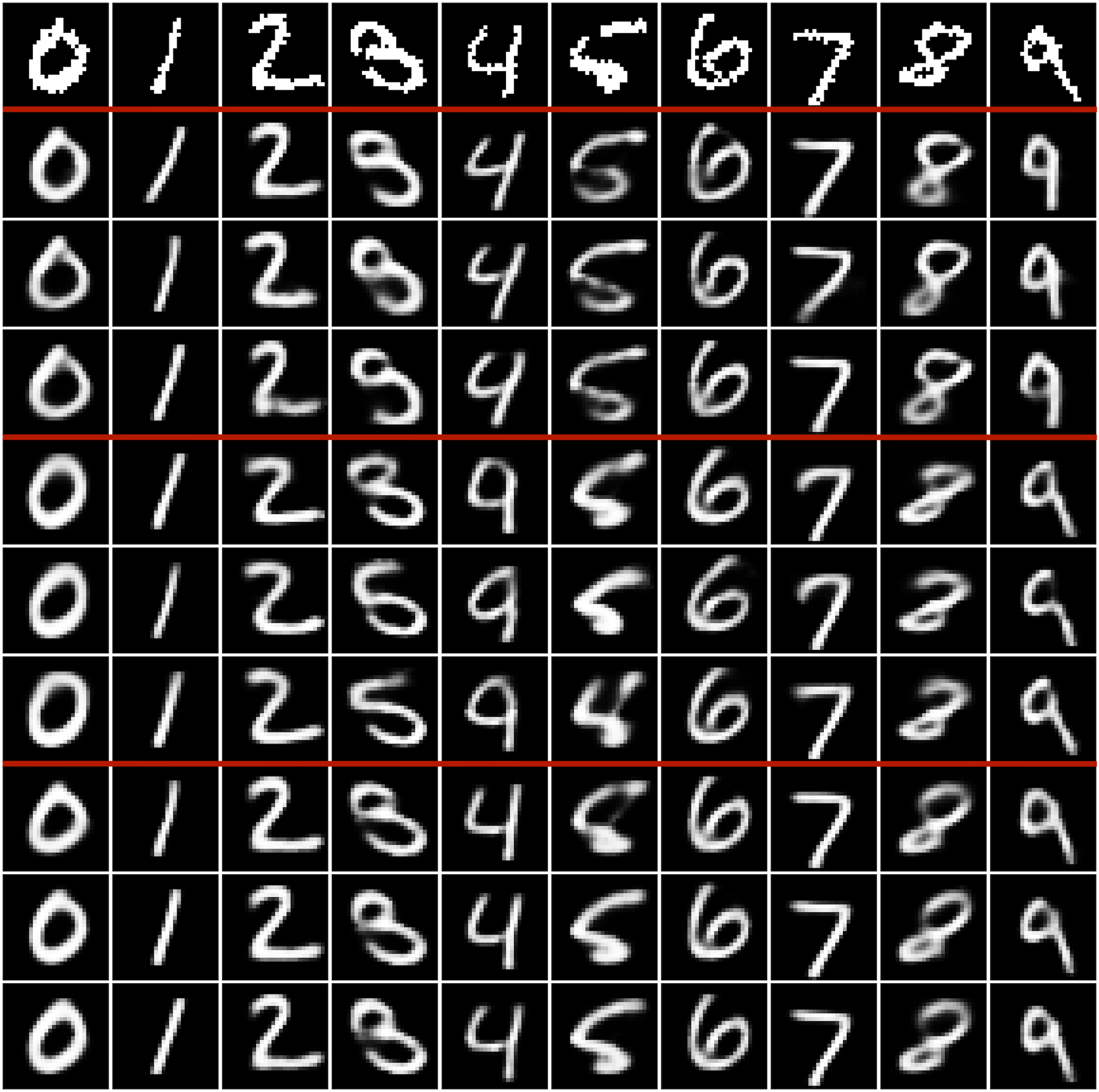}
	\caption{\textbf{Row 1:} Original images. 
	\textbf{Row 2--4:} Belief informed by top half of the image. \textbf{Row 5--7:} Informed by bottom half. 			    \textbf{Row 8--10:} Integrated belief.}
   	\label{fig:mnist_digits_top_bot_int}
   \end{subfigure}
   \quad
	\begin{subfigure}[t]{0.47\columnwidth} \centering
		\includegraphics[width=1.0\columnwidth]{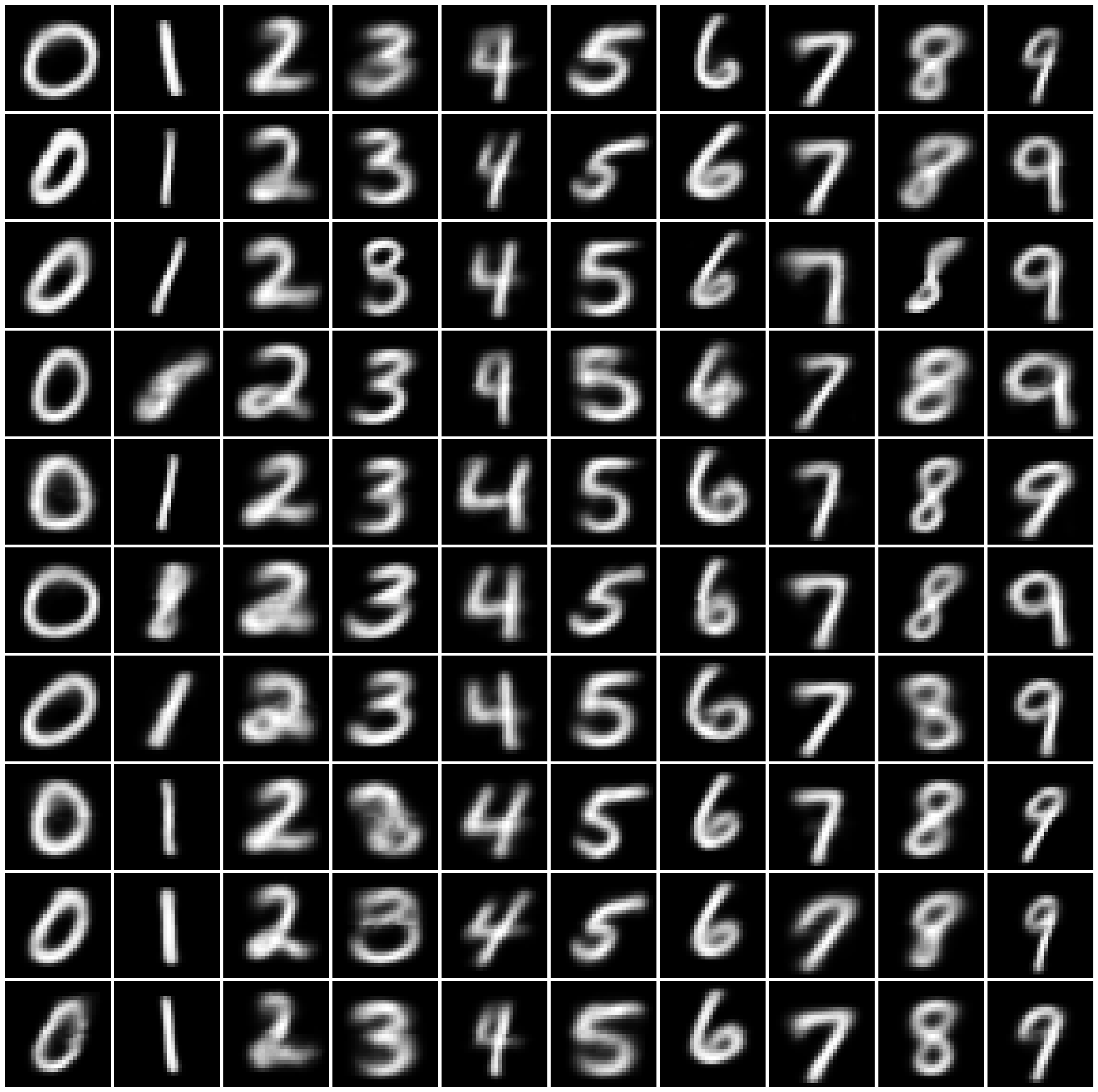}
		\caption{Predictions from 10 random samples of the latent variables, inferred from one-hot class labels.}
		\label{fig:mnist_digits_from_labels}
	\end{subfigure}
\caption{Predicted images, where latent variables are inferred from the variational distributions of different sources. 
Sources with partial information generate diverse samples, the integration resolves ambiguities. 
E.g.~in Fig.~\ref{fig:mnist_digits_top_bot_int}, the lower half of digit 3 randomly generates digits 5 and 3 and the upper half generates digits 3 and 9. In contrast, the integration resolves ambiguities.}
\end{figure}

Next, we evaluate our method for conditional (structured) prediction using Eq.~(\ref{eq:conditional_prediction}). 
Fig.~\ref{fig:mnist_digits_top_bot_int} shows the means of the likelihood functions, with latent variables drawn from individual and integrated beliefs. 
To demonstrate conditional image generation from labels, we add a third encoder that perceives class labels. 
Fig.~\ref{fig:mnist_digits_from_labels} shows the means of the likelihood functions, inferred from labels. 

\begin{figure}
\centering
\begin{subfigure}[t]{0.47\columnwidth}
   \centering
	\includegraphics[width=1.0\columnwidth]{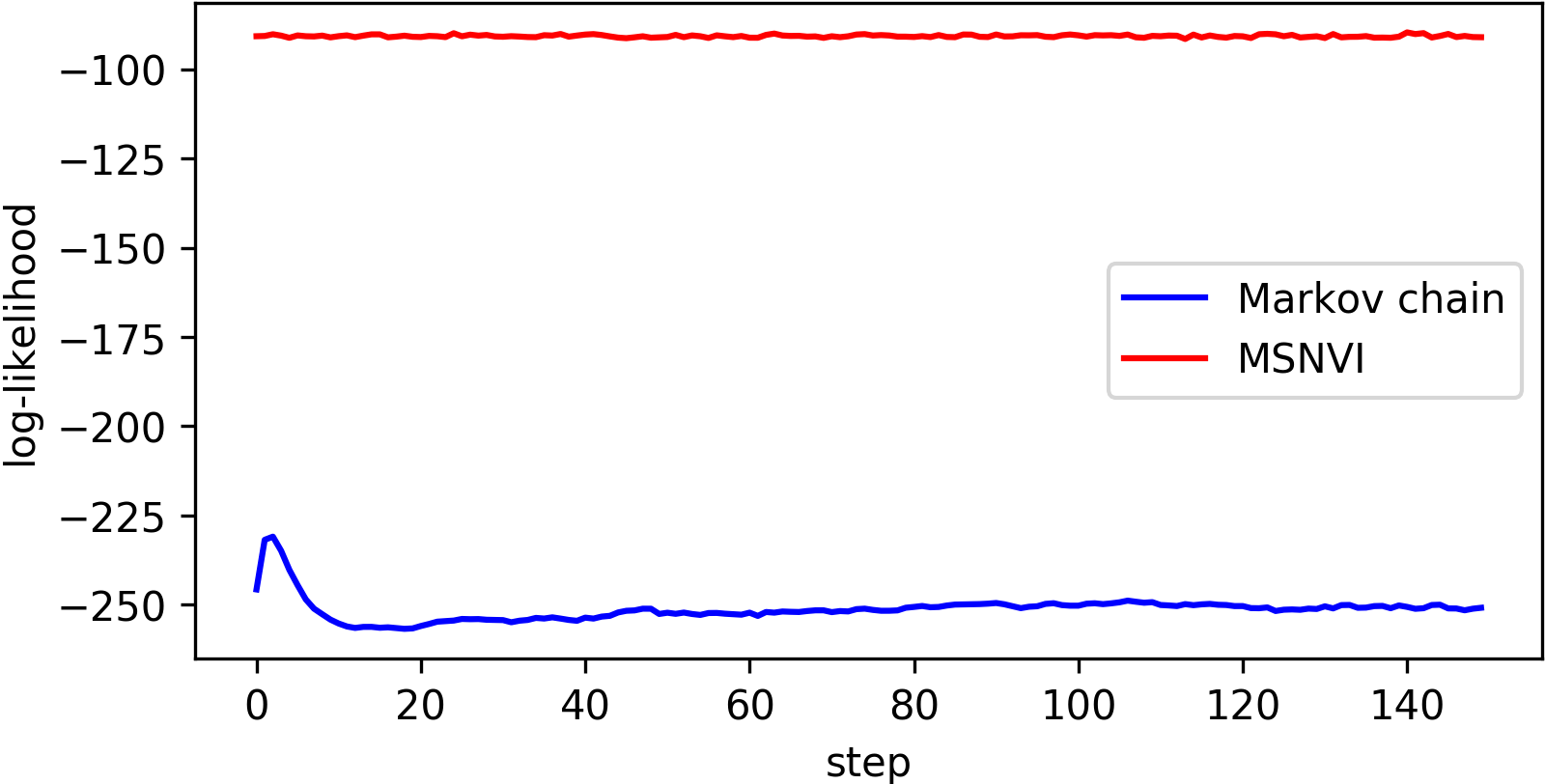}
	\caption{MNIST-TB, where bottom half is missing.}
\label{fig:missing_data_inference_marginal_mnist_tb}
\end{subfigure}
\quad
\begin{subfigure}[t]{0.47\columnwidth}
	\centering
	\includegraphics[width=1.0\columnwidth]{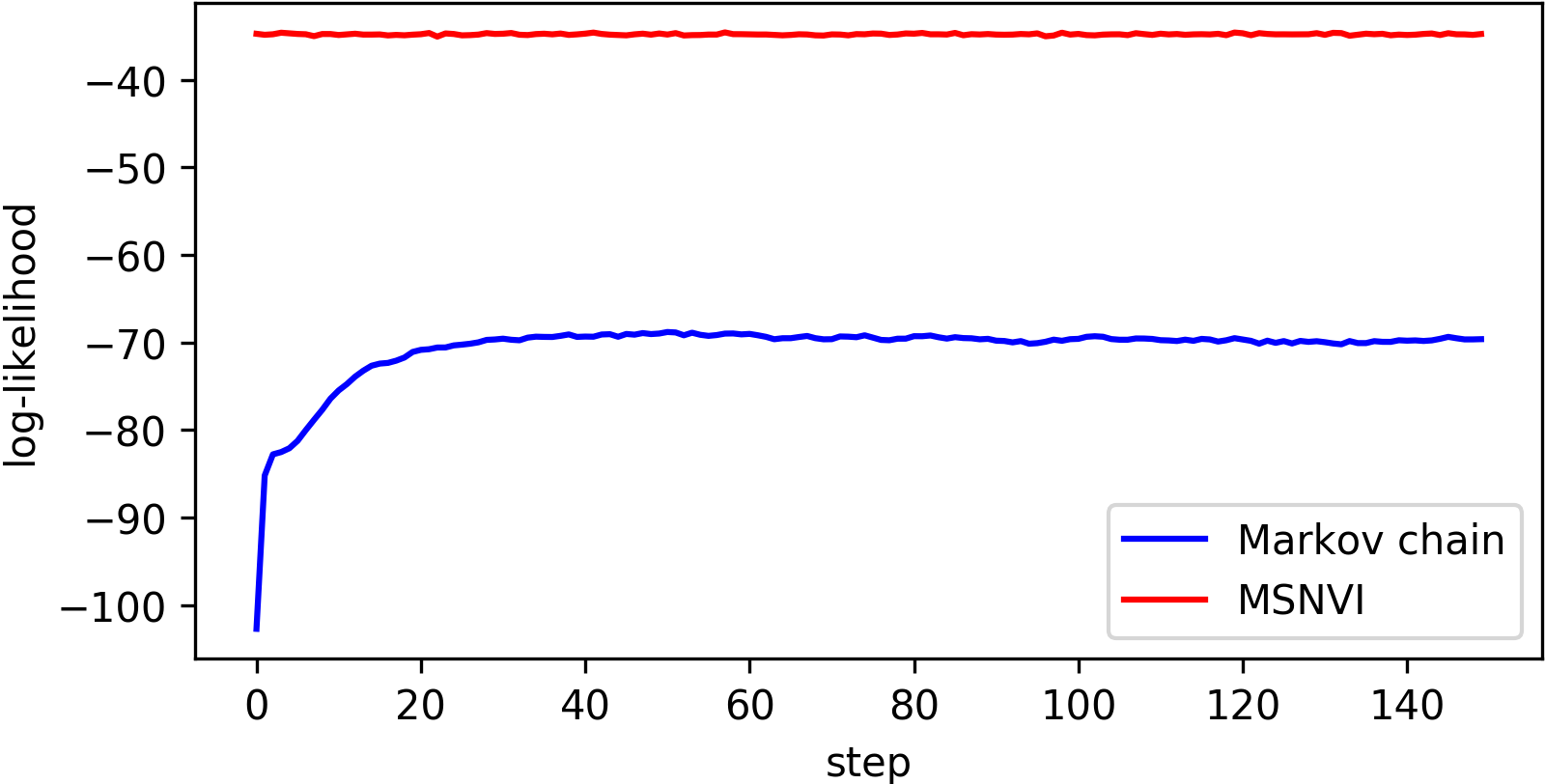}
	\caption{MNIST-QU, where bottom right quarter is missing.}	\label{fig:missing_data_inference_marginal_mnist_qu}
\end{subfigure}
\caption{Missing data imputation with Monte Carlo procedure described in \cite{RezendeMW14} and our method. 
For the Markov chain procedure, the initial missing data is drawn randomly from $\operatorname{Ber}\left({0.5}\right)$ and imputed from the previous random generation in subsequent steps. 
MSNVI was trained with $\elboa$. 
For MNIST-QU, we used the PoE belief of the three observed quarters.
The plots show the log-likelihood at every step of the Markov chain, marginalised over the dataset. 
Higher is better.
}
\label{fig:missing_data_inference_marginal_mnist}
\end{figure}

We also compare our method to the missing data imputation procedure described in \cite{RezendeMW14} for MNIST-TB und MNIST-QU. 
We run the Markov chain for all samples in the test set for 150 steps each and calculate the log likelihood of the imputed data at every step. 
The results---averaged over the dataset---are compared to our multimodal data generation method in Fig.~\ref{fig:missing_data_inference_marginal_mnist}. 
For large portions of missing data as in MNIST-TB, the Markov chain often fails to converge to the marginal distribution. 
But even for MNIST-QU with only a quarter of the image missing, our method outperforms the Markov chain procedure by a large margin. 
Please consult the supplementary material for a visualisation of the stepwise generations during the inference procedure.
\subsubsection{Caltech-UCSD Birds 200}
Caltech-UCSD Birds 200 \cite{WelinderEtal2010} is a dataset with 6033 images of birds with $128\times128$ resolutions, split into 3000 train and 3033 test images. 
As a second source, we use segmentation masks provided by \cite{Yang2014MaxMarginBM}. 
On this dataset we assess whether learning with multiple modalities can be advantageous in scenarios where we are interested only in one particular modality. 
Therefore, we evaluate the ELBO for a single source and a single target observation, i.e.~encoding images and decoding segmentation masks. 
We compare models that learned with multiple modalities using $\elboa$ and $\elboac$ with models that learnt from a single modality. 
Additionally, we evaluate the segmentation accuracy using Eq.~(\ref{eq:conditional_prediction}). 
The accuracy is estimated with 100 samples, drawn from the belief informed by image data. 
The results are summarised in Tab.~\ref{tab:cub_elbo_and_accuracy}. 
\begin{table}
\caption{Negative ELBOs and segmentation accuracy on Caltech-UCSD Birds 200. 
The IWAE was trained with a single source and target observation. 
Models trained with $\elboa$ and $\elboac$ use all sources and targets, and $\elboa$* and $\elboac$* use all sources for inference, but learn the generative model of a single modality.}
\resizebox{1.0\columnwidth}{!} {
\begin{tabular}{llllll}
\toprule
& $\elboa$ & $\elboa$* & $\elboac$ & $\elboac$* & IWAE \\
\midrule
img-to-seg   & 5326   	& 3264     & 5924	  & 3337	& \textbf{3228}  \\
img-to-img	 & -26179 & -26663 & -29285 & -29668	& \textbf{-30415}  \\
accuracy     & 0.808 & 0.870 & 0.810 & \textbf{0.872}  & 0.855	\\
\bottomrule
\end{tabular}
}
\label{tab:cub_elbo_and_accuracy}
\end{table}
We distinguish between objectives that involve both modalities in the generative model and objectives where we learn only the generative model for the modality of interest (segmentation), denoted with an asterisk.
Models that have to learn the generative models for images and segmentations show worse ELBOs and accuracy, when evaluated on one modality. 
In contrast, the accuracy is slightly increased when we learn the generative model of segmentations only, but use both sources for inference. 
\newline
We also refer the reader to the supplementary material, where we visualise conditionally generated images, showing that learning with the importance sampling estimate of the ELBO is crucial to generate diverse samples from partially informed sources. 
\subsection{Robustness via conflict detection and redundancy}
\begin{figure*}[ht!]
\centering
\begin{subfigure}[t]{0.31\textwidth}
	\hfill
	\includegraphics[width=0.85\textwidth]{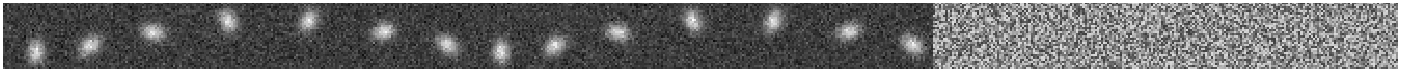}
	\label{fig:pendulum_imgs_sensor_0}
\end{subfigure}
\quad
\begin{subfigure}[t]{0.31\textwidth}
	\hfill
	\includegraphics[width=0.85\textwidth]{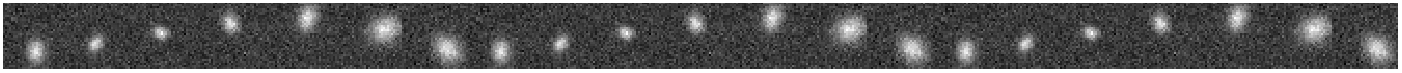}
	\label{fig:pendulum_imgs_sensor_1}
\end{subfigure}
\quad
\begin{subfigure}[t]{0.31\textwidth}
	\hfill
	\includegraphics[width=0.85\textwidth]{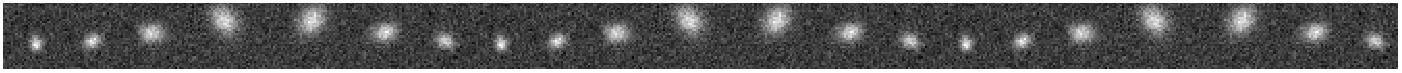}
	\label{fig:pendulum_imgs_sensor_2}
\end{subfigure}
\begin{subfigure}[t]{0.31\textwidth}
	\centering
	\includegraphics[width=1.0\textwidth]{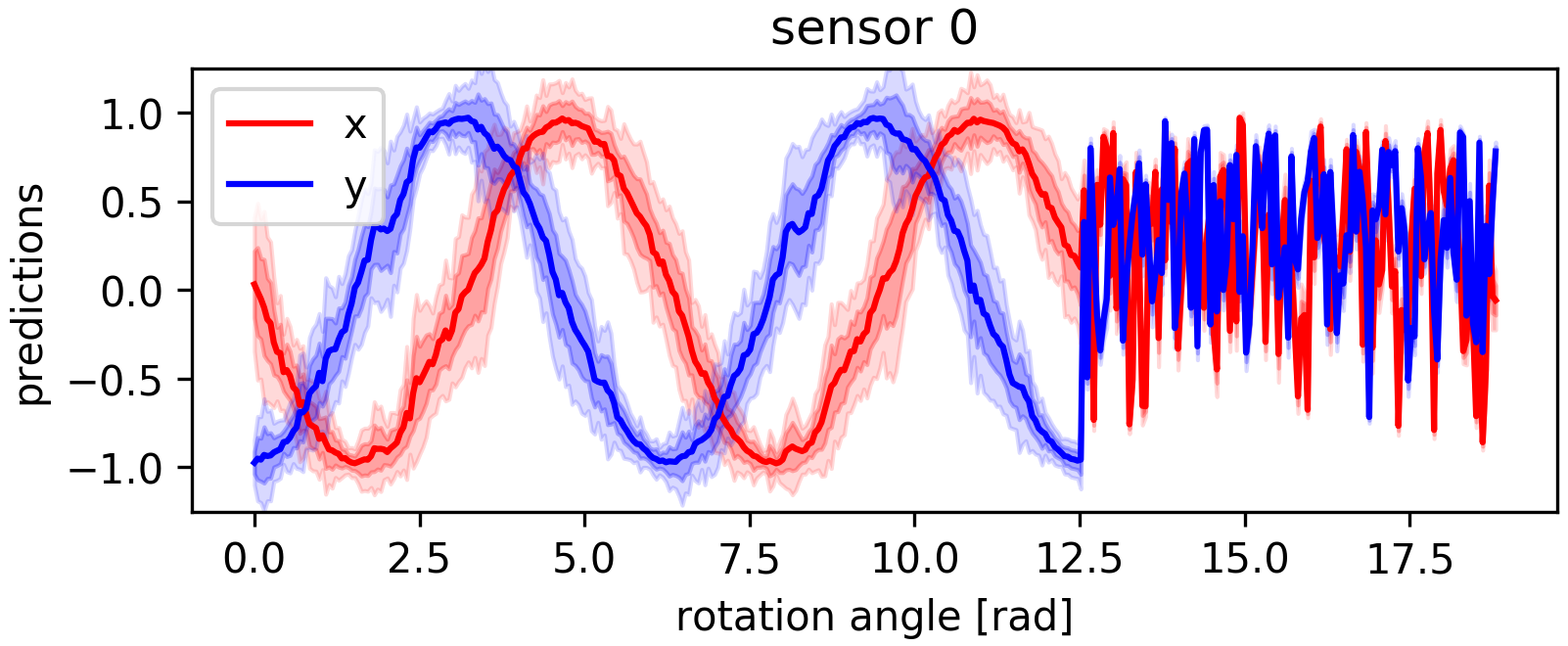}
	\label{fig:pendulum_robustness_sensor_0}
\end{subfigure}
\quad
\begin{subfigure}[t]{0.31\textwidth}
	\centering
	\includegraphics[width=1.0\textwidth]{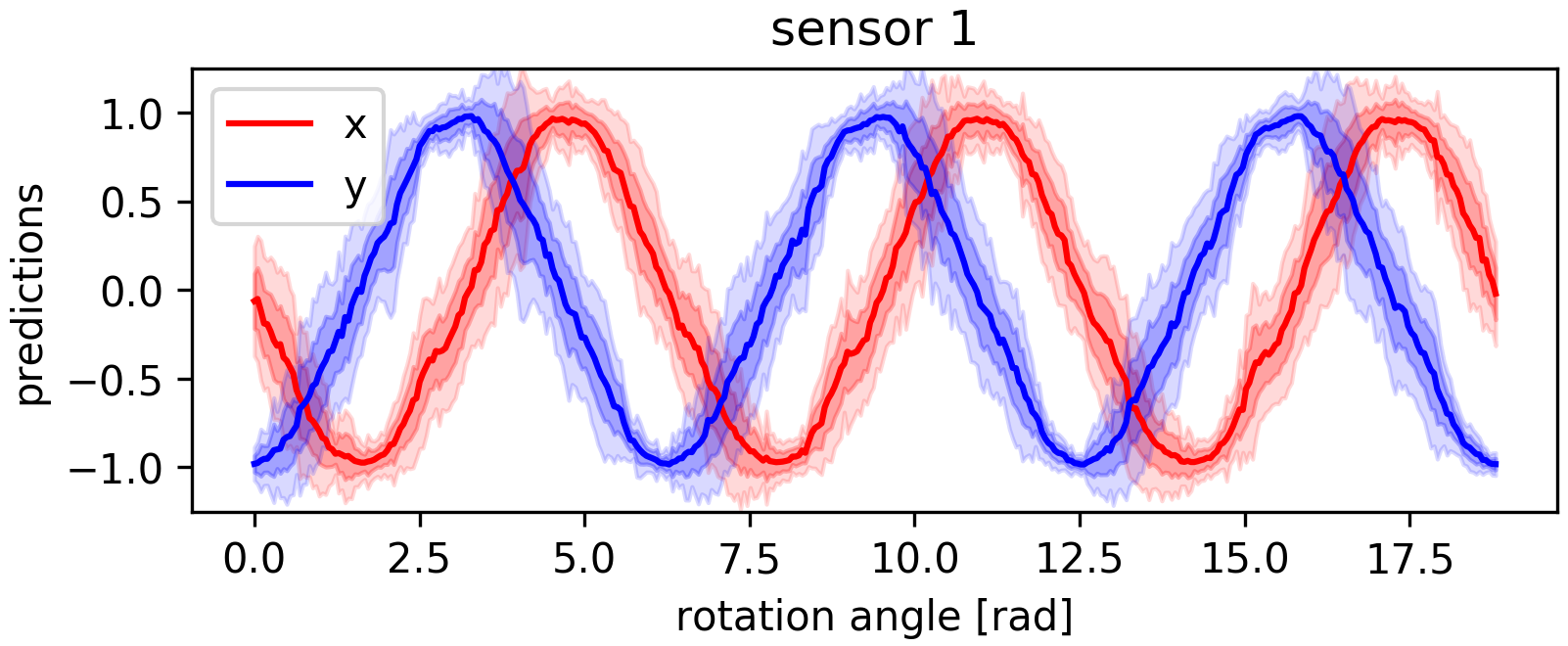}
	\label{fig:pendulum_robustness_sensor_1}
\end{subfigure}
\quad
\begin{subfigure}[t]{0.31\textwidth}
	\centering
	\includegraphics[width=1.0\textwidth]{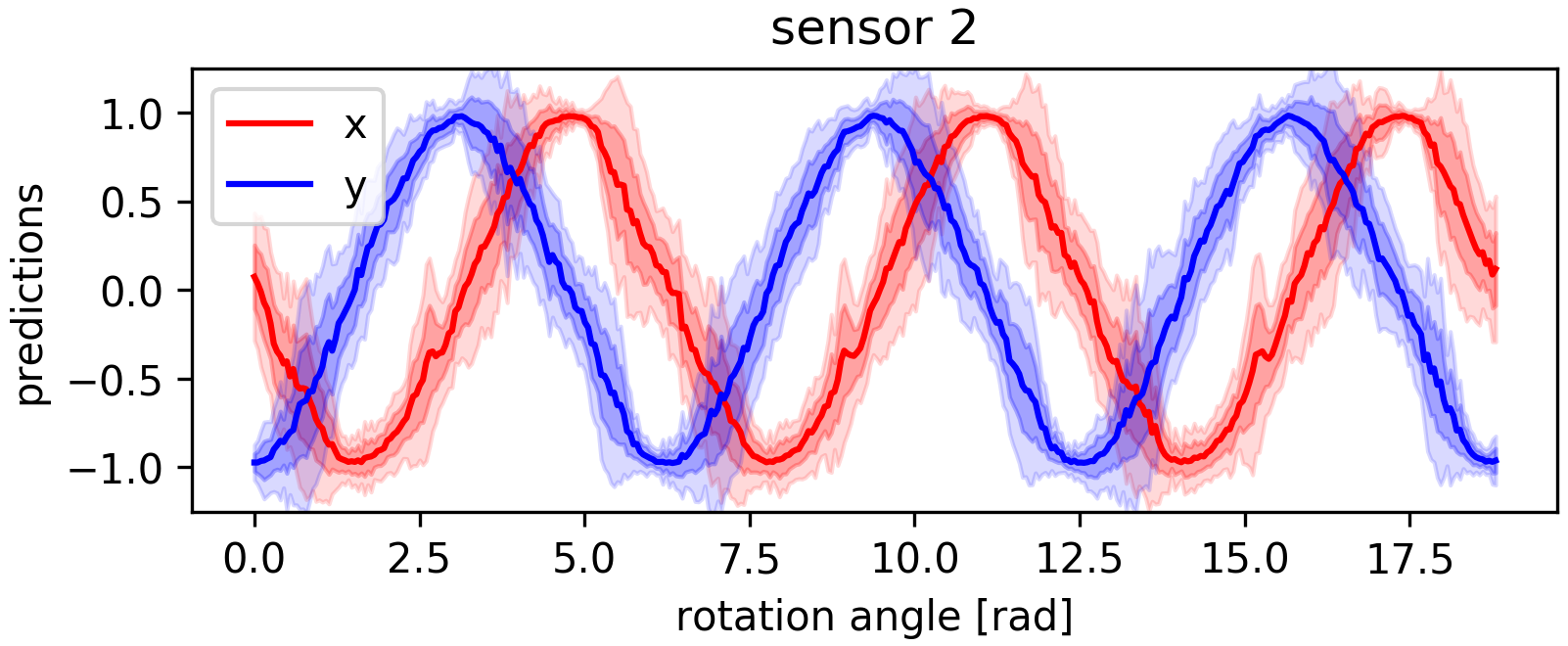}
	\label{fig:pendulum_robustness_sensor_2}
\end{subfigure}
\quad
\begin{subfigure}[t]{0.31\textwidth}
	\centering
	\includegraphics[width=1.0\textwidth]{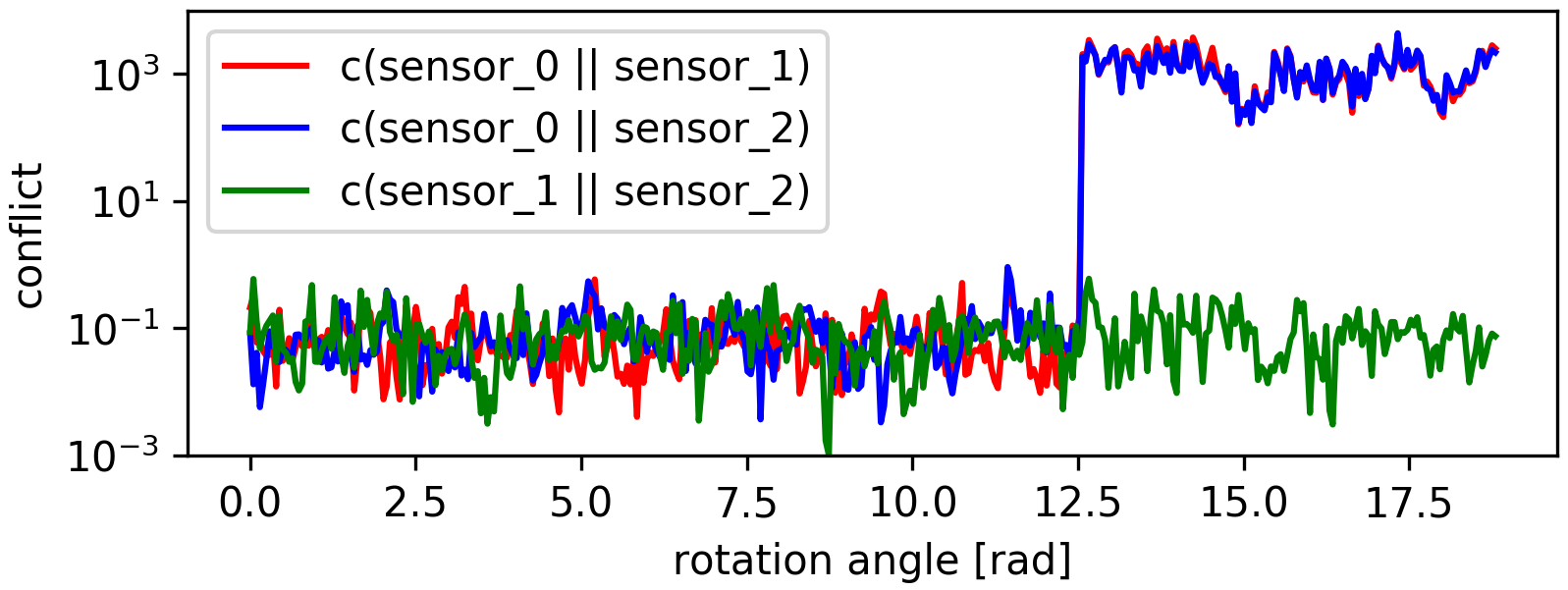}
	\label{fig:pendulum_robustness_conflict}
\end{subfigure}
\quad
\begin{subfigure}[t]{0.31\textwidth}
	\centering
	\includegraphics[width=1.0\textwidth]{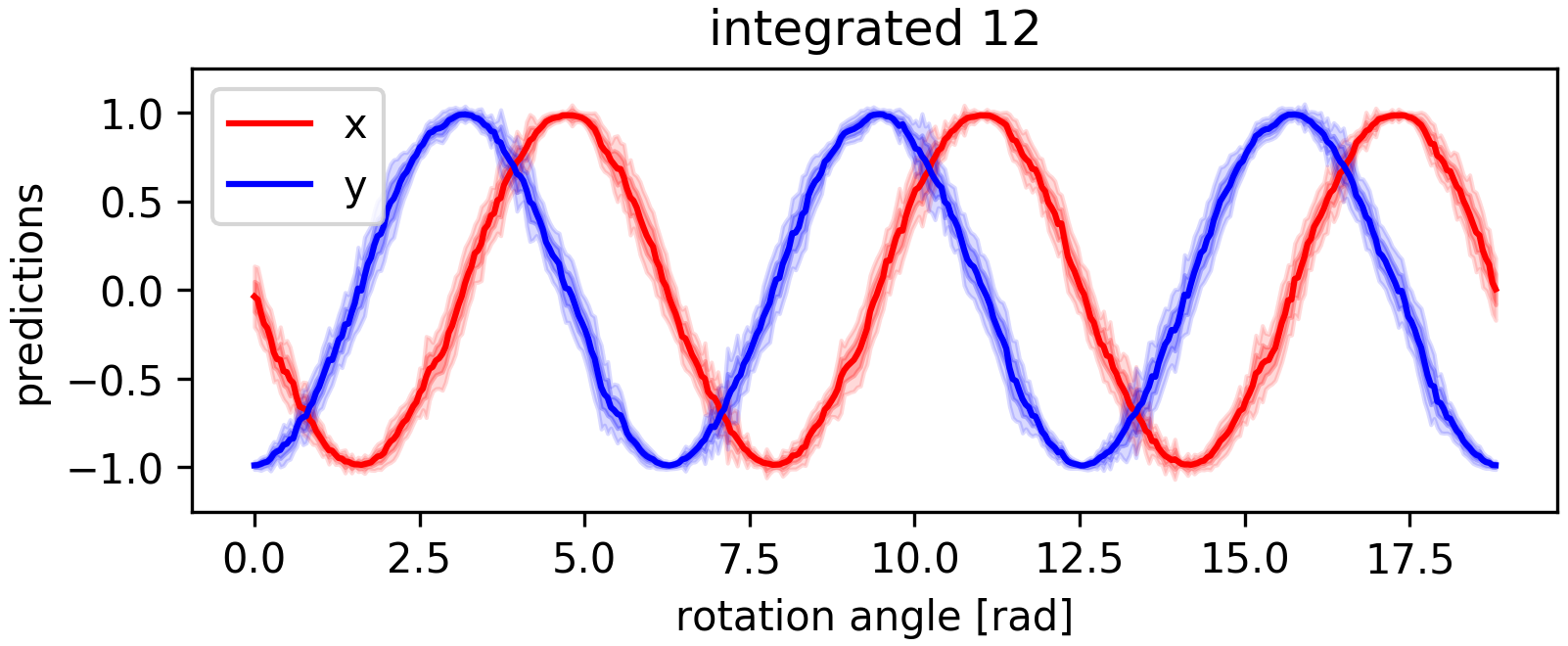}
	\label{fig:pendulum_robustness_sensor_12}
\end{subfigure}
\quad
\begin{subfigure}[t]{0.31\textwidth}
	\centering
	\includegraphics[width=1.0\textwidth]{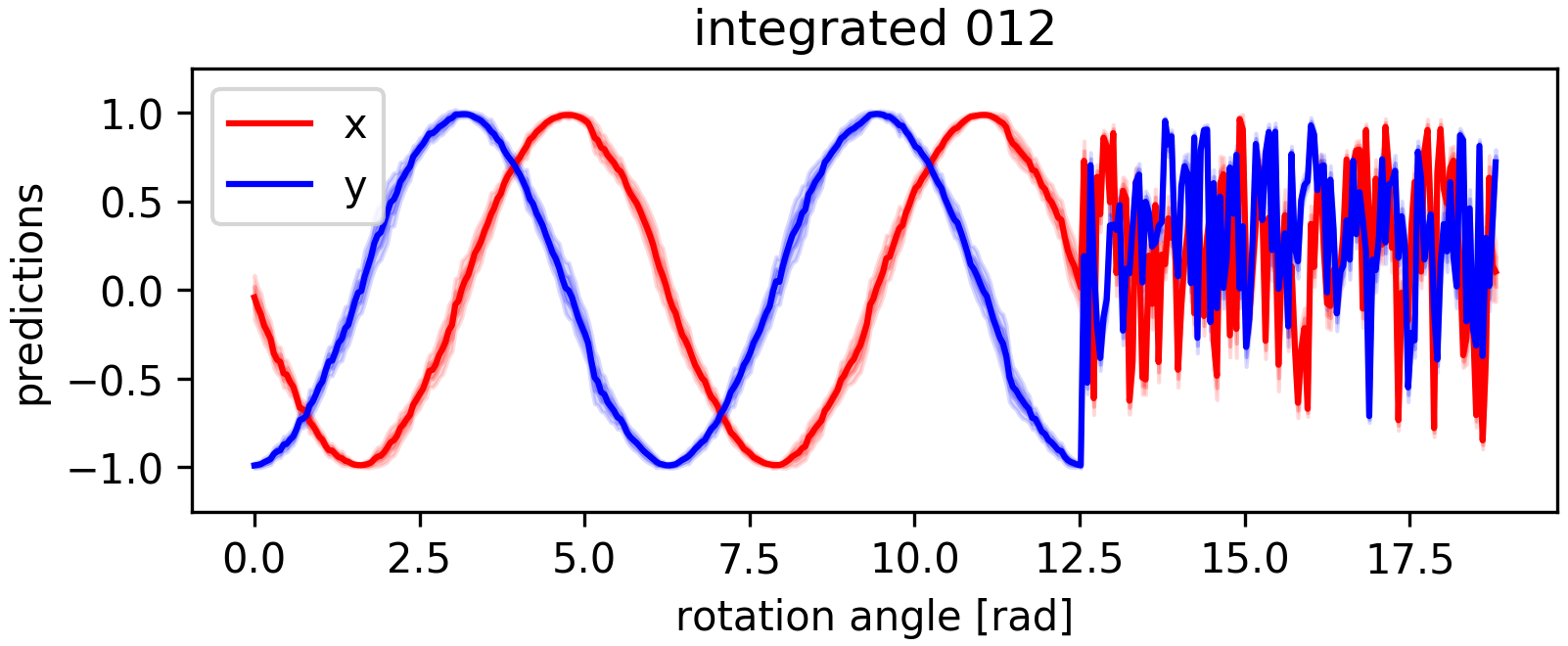}
	\label{fig:pendulum_robustness_sensor_012}
\end{subfigure}
\caption{Predictions ($x$- and $y$-coordinates) of the pendulum position (figures 1, 2, 3, 5, 6) and conflict measure (figure 4). For the predictions, latent variables are inferred from images of 3 sensors with different views (top row) as well as their integrated beliefs (bottom mid and right). 
The figures show predictions (of the static model) for different angles of the pendulum, performing 3 rotations.
After 2 rotations, failure of sensor 0 is simulated by outputting noise only.
Lines show the mean and shaded areas show 1 and 2 standard deviations, estimated using 500 random samples of latent variables.
Bottom left: The conflict measure of Eq.~(\ref{eq:conflict}) for different angles of the pendulum. 
}  
\label{fig:pendulum_robustness}
\end{figure*}
In this experiment we demonstrate how a shared latent representation can increase robustness, by exploiting sensor redundancy and the ability to detect conflicting data. 
We created a synthetic dataset of perspective images of a pendulum with different views of the same scene. 
The pendulum rotates along the z-axis and is centred at the origin. 
We simulate three cameras with $32\times32$-pixel resolution as information sources for inference and apply independent noise with std $0.1$ to all sources. 
Each sensor is directed towards the origin (centre of rotation) from different view-points: Sensor 0 is aligned with the $z$-axis, and sensor 1 and 2 are rotated by $45\deg$ along the $x$- and $y$-axis, respectively.
The distance of all sensors to the origin is twice the radius of the pendulum rotation. 
For the generative model we use the $x$- and $y$-coordinate of the pendulum rather than reconstructing the images. 
The model was trained with $\elboa$. 

In Fig.~\ref{fig:pendulum_robustness}, we plot the mean and standard deviation of predicted $x$- and $y$-coordinates, where latent variables are inferred from a single source as well as from the PoE posteriors of different subsets. 
As expected, integrating the beliefs from redundant sensors reduces the predictive uncertainty.
Additionally, we visualise the three images used as information sources above these plots. 

Next, we simulate an anomaly in the form of a defect sensor 0, outputting random noise after 2 rotations of the pendulum. 
This has a detrimental effect on the integrated beliefs, where sensor 0 is part of the integration.
We also plot the conflict measure of Eq.~(\ref{eq:conflict}). 
As can be seen, the conflict measures for sensor 0 increases significantly when sensor 0 fails.
In this case, one should integrate only the two remaining sensors with low conflict conjunctively. 
\section{Summary and future research directions}\label{sec:conclusion}
We extended neural variational inference to scenarios where multiple information sources are available. 
We proposed an objective function to learn individual inference models jointly with a shared generative model. 
We defined an exemplar measure (of conflict) to compare the beliefs from distinct inference models and their respective information sources.
Furthermore, we proposed a disjunctive and a conjunctive integration method to combine arbitrary subsets of beliefs.

We compared the proposed objective functions experimentally, highlighting the advantages and drawbacks of each. 
Naive integration as a PoE ($\elboc$) leads to inseparable individual beliefs, while optimising the sources only individually ($\elboa$) worsens the integration of the sources.
On the other hand, a hybrid of the two objectives ($\elboac$) achieves a good trade-off between both desiderata. 
Moreover, we showed how our method can be applied to structured output prediction and the benefits of exploiting the comparability of beliefs to increase robustness. 

This work offers several future research directions. 
As an initial step, we considered only static data and a simple latent variable model. 
However, we have made no assumptions about the type of information source. 
Interesting research directions are extensions to sequence models, hierarchical models and different forms of information sources such as external memory. 
Another important research direction is the combination of disjunctive and conjunctive integration methods, taking into account the conflict between sources. 
\section*{Acknowledgements} 
We would like to thank Botond Cseke for valuable suggestions and discussions.
\small{
\bibliography{msnvi}
}
\bibliographystyle{aaai}
\renewcommand\thesubsection{\Alph{subsection}}
\clearpage
\section{Appendix}
\subsection{Individual inferences} \label{sec:appendix:boundA}
In this section we derive the $\elboa$. 
Since any proposal distribution yields an ELBO to the log-marginal likelihood, the (weighted) average is also an ELBO.
\begin{equation*} 
\begin{aligned}
& \ln p_{\theta}(\mathbf{X}) = \sum_{n=1}^{N} \ln p_{\theta}(\mathbf{x}^{(n)})
\end{aligned}
\end{equation*}
\begin{equation*}
\begin{aligned}
\ln p_{\theta}(\mathbf{x}^{(n)}) &=  \ln 
\sum_{m=1}^{M} \pi_{m}
\mathbb{E}_{\z^{(n)}_{1:K} \sim q_{\phi_m} \big(\z^{(n)} \given \x_{m}^{(n)}\big)} \Big{[} 
\frac{1}{K} \sum_{k=1}^{K}
w^{(n)}_{m, k}
\Big{]}  \\
& \geq \sum_{m=1}^{M} \pi_{m} \ln 
\mathbb{E}_{\z^{(n)}_{1:K} \sim q_{\phi_m} \big(\z^{(n)} \given \x_{m}^{(n)}\big)} \Big{[} 
\frac{1}{K} \sum_{k=1}^{K}
w^{(n)}_{m, k}
\Big{]}  \\
& \geq \sum_{m=1}^{M} \pi_{m} \mathbb{E}_{\z^{(n)}_{1:K} \sim q_{\phi_m} \big(\z^{(n)} \given \x_{m}^{(n)}\big)} \Big{[} 
\ln \frac{1}{K} \sum_{k=1}^{K} w^{(n)}_{m, k} \Big{]}, 
\end{aligned}
\end{equation*}
where 
\begin{equation*}
w^{(n)}_{m, k} = \frac{p_{\theta}(\mathbf{x}^{(n)} \given \mathbf{z}^{(n)}_{k}) \, p_{\theta}(\mathbf{z}^{(n)}_{k})}{q_{\phi_m}(\z_{k}^{(n)} \given \x_{m}^{(n)})}.
\end{equation*}
The factors $\pi_{m}$ are the weights for each ELBO term, satisfying $0 \leq \pi_{m} \leq 1$ and $\sum_{m=1}^{M}\pi_{m} = 1$. 
\newline
When $K=1$, the gap between $\elboa$ and the marginal log-likelihood is the average Kullback-Leibler (KL) divergence between individual approximate posteriors and the true posterior from all sources:
\begin{equation*} \label{eq:boundA_gap}
\begin{aligned}
& \ln p_{\theta}(\x^{(n)}) - \elboa \\
&= \sum_{m=1}^{M} \pi_{m} D_{\mathrm{KL}}\big(q_{\phi_{m}}(\z^{(n)} \given \x^{(n)}_{m}) ~||~ p_{\alltheta}(\z^{(n)} \given \allx^{(n)} )\big)
\end{aligned}
\end{equation*}
This gap can be further decomposed as:
\begin{equation*}
\begin{aligned}
&\sum_{m=1}^{M} \pi_{m} \mathrm{KL}(q_{\phi_{m}}(\z \given \x_{m}) ~||~ p(\z \given \x)) \\
&= \sum_{m=1}^{M} \pi_{m} \mathrm{KL}(q_{\phi_{m}}(\z \given \x_{m}) ~||~ p(\z \given \x_{m})) \\
&- \sum_{m=1}^{M} \pi_{m} \mathbb{E}_{\z \sim q_{\theta_{m}}(\z \given \x_{m})} 
\big[\ln p_{\theta_{-m}}(\x_{-m} \given \z) \big] \\
&+ \sum_{m=1}^{M} \pi_{m}\ln p_{\theta_{-m}}(\x_{-m} \given \x_{m}) \\
&= \sum_{m=1}^{M} \pi_{m} \mathrm{KL}(q_{\phi_{m}}(\z \given \x_{m}) ~||~ p(\z \given \x_{m})) \\
&- \sum_{m=1}^{M} \pi_{m} \mathbb{E}_{\z \sim q_{\theta_{m}}(\z \given \x_{m})} 
\big[\ln p_{\theta_{-m}}(\x_{-m} \given \z) \big] \\
&+ \sum_{m=1}^{M} \pi_{m}\ln \mathbb{E}_{\z \sim p(\z \given \x_{m})}  \big[ p_{\theta_{-m}}(\x_{-m} \given \z) \big].
\end{aligned}
\end{equation*}
To minimise $\elboa$, not only the KL divergence of the individual approximate posterior and the respective true posterior need to be minimised, but also two additional terms which depend on the likelihood of those observations that have not been used as an information source for inference. 
%
%
%
\subsection{Mixture of experts inference} \label{sec:appendix:boundB}
The ELBO for the mixture distribution $\elbob$ can be derived similarly. 
We employ a Monte Carlo approximation only w.r.t. each mixture component but not w.r.t. the mixture weights. 
That is, we enumerate all possible mixture components rather than sampling each from an indicator variable. 
This reduces variance of the estimate and circumvents the problem of propagating gradients through the sampling process of discrete random variables. 
\begin{equation*}
\begin{aligned}
& \ln p_{\theta}\big(\x^{(n)}\big) \\
&= \sum_{n=1}^{N} \ln 
\int 
p_{\theta}\big(\x^{(n)}, \z^{(n)}\big) \frac{\sum_{m=1}^{M} \pi_{m} q_{\phi_m}(\z^{(n)} \given \x_{m}^{(n)})}{\sum_{m'=1}^{M} \pi_{m'} q_{\phi_{m'}}(\z^{(n)} \given \x_{m'}^{(n)})} d\z^{(n)} \\
& = \ln \sum_{m=1}^{M} \pi_{m} 
\mathbb{E}_{\z^{(n)} \sim q_{\phi_m}\big(\z^{(n)} \given \x_{m}^{(n)}\big)} 
\Bigg{[} 
w^{(n)}_{m, k} 
\Bigg{]}  \\
& = \ln \sum_{m=1}^{M} \pi_{m} 
\mathbb{E}_{\z^{(n)} \sim q_{\phi_m}\big(\z^{(n)} \given \x_{m}^{(n)}\big)} 
\Bigg{[} 
\frac{1}{K} \sum_{k=1}^{K} w^{(n)}_{m, k} 
\Bigg{]}  \\
& \geq \sum_{m=1}^{M} \pi_{m} \ln 
\mathbb{E}_{\z^{(n)} \sim q_{\phi_m}\big(\z^{(n)} \given \x_{m}^{(n)}\big)} 
\Bigg{[} 
\frac{1}{K} \sum_{k=1}^{K} w^{(n)}_{m, k} 
\Bigg{]}  \\
& \geq \sum_{m=1}^{M} \pi_{m} \mathbb{E}_{\z^{(n)}_{1:K} \sim q_{\phi_m} \big(\z^{(n)} \given \x_{m}^{(n)}\big)} \Bigg{[} 
\ln \frac{1}{K} \sum_{k=1}^{K} w^{(n)}_{m, k} 
\Bigg{]}, 
\end{aligned}
\end{equation*}
$\elbob$ minimises the average KL-divergence between the mixture of approximate posteriors and the true posterior from all sources:
\begin{equation*} \label{eq:boundB_gap}
\begin{aligned}
& \ln p(\x) - \elbob \\
&=  \sum_{m=1}^{M} \pi_{m} D_{\mathrm{KL}} \Big(\sum_{m'=1}^{M} \pi_{m'} q_{\phi_{m'}}(\z \given \x_{m'}) ~||~ p(\z \given \x)\Big).
\end{aligned}
\end{equation*}
\subsection{Product of Gaussian experts} \label{sec:appendix:boundC_gaussian}
Here we consider the popular case of individual Gaussian approximate posteriors and a zero-centred Gaussian prior.  
Let the normal distributions be represented in the canonical form with canonical parameters $\{ \Lambda; \mathbf{\eta}\}$:
\begin{equation*}
p(\z) = \frac{1}{Z(\mu, \Lambda)} \exp \Big( \mathbf{\eta}^{T} \z - \frac{1}{2} \z^{T} \Lambda \z \Big).
\end{equation*}
$\Lambda$ denotes the precision matrix and $\eta = \Lambda \mu$, where $\mu$ is the mean. 
Furthermore, $Z(\mu, \Lambda)=(2 \pi)^{D_{z}/2} | \Lambda|^{-1/2} \exp(\frac{1}{2} \eta^{T} \Lambda^{-1} \eta)$ is the partition function.

Let the subscripts $m$, $0$ and $*$ indicate the $m$-th approximate distribution, the prior and the integrated distribution. 
The natural parameters of the integrated variational posterior $q_{\allphi}(\z  \allx)$ from Eq.~(\ref{eq:posterior_product}) can then be calculated as follows: 
\begin{equation*}
\Lambda_{*} = \sum_{m=1}^{M} \Lambda_{m} - (M-1) \Lambda_{0},
\end{equation*}
\begin{equation*}
\eta_{*} = \sum_{m=1}^{M} \eta_{m} - (M-1) \eta_{0}.
\end{equation*}
To obtain a valid integrated variational posterior, we require the precision matrix $\Lambda_{*}$ to be positive semi-definite. 
This enforces requirements for the precision matrices $\Lambda_{m}$.
In the case of diagonal precision matrices, the necessary and sufficient condition is that $\Lambda_{*}$ has all positive entries. 
A sufficient condition for each entry $\Lambda_{m}[i]$ is $\Lambda_{0}[i] \leq \Lambda_{m}[i]$. 

The partition function of the integrated belief can be calculated from the natural parameters, taking $\eta_{*} = \Lambda_{*} \mu_{*}$:
\begin{equation}\label{eq:partition_function_from_int_params}
Z(\mu_{*}, \Lambda_{*}) = (2 \pi)^{D_{z}/2} |\Lambda_{*}|^{-1/2} \exp \Big( \frac{1}{2} (\eta_{*})^{T} (\Lambda_{*})^{-1} \eta_{*} \Big).
\end{equation}
\subsection{Point-wise mutual information}
Inspecting Eq.~(\ref{eq:posterior_product}), we can see that the negative logarithm of the constant term corresponds to the pointwise mutual information (PMI) between the observations. 
We do not need to calculate this constant since we impose assumptions about the parametric forms of the distributions and can calculate the partition function $Z(\mu_{*}, \Lambda_{*})$ of the integrated belief using Eq.~(\ref{eq:partition_function_from_int_params}). 

However, we can also calculate this partition function from the product of individual partition functions and the above mentioned constant in Eq.~(\ref{eq:posterior_product}):
\begin{equation*}
\begin{aligned}
&\frac{1}{Z(\mu_{*}, \Lambda_{*})} \\
&= \left[ \prod_{m=1}^{M} \frac{1}{Z(\mu_{m}, \Lambda_{m})} \right] \cdot 
Z(\mu_{0}, \Lambda_{0})^{(M-1)} \cdot
\frac{\prod_{m=1}^{M} p(\x_{m})}{p(\allx)}.
\end{aligned}
\end{equation*}
The PMI can then be calculated as: 
\begin{equation*} \label{eq:pmi}
\mathrm{PMI}  = \ln \bigg(
\frac{Z(\mu_{0}, \Lambda_{0}) ^{M-1}} {\prod_{m=1}^{M} Z(\mu_{m}, \Lambda_{m})} \cdot Z(\mu_{*}, \Lambda_{*}) \bigg).
\end{equation*}
The pointwise mutual information can be calculated between any subset of information sources. 
We note however, that it is based on the assumption that all involved probability density functions---the prior and all approximate posterior distributions---are normal distributions. 
\subsection{Visualisation of samples from individual and integrated beliefs on mixture of bi-variate Gaussians dataset}
\begin{figure}[h!]\centering
\caption{Samples from individual and integrated beliefs and samples obtained after SIR}
\begin{subfigure}[t]{0.96\columnwidth} \centering
\includegraphics[width=1.0\columnwidth]{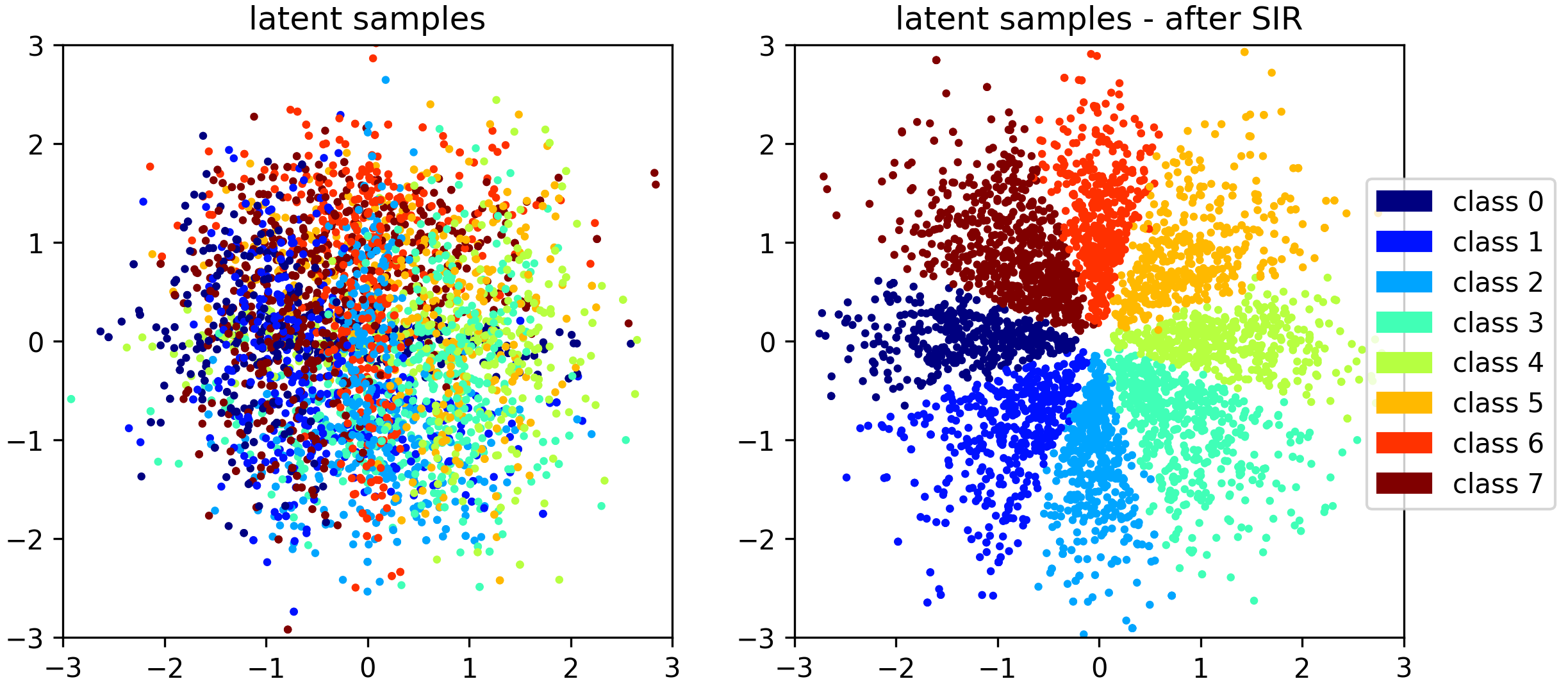}
\caption{Individual beliefs and their predictions.
\textbf{Left:} Random samples from variational posterior without integration. Colours correspond to 8 test points, located at the means of the mixture of Gaussians data distribution. 
\textbf{Right:} Samples after sampling importance re-sampling using all likelihood functions.
}
\label{fig:samples_and_SIR_individual}
\end{subfigure}

\begin{subfigure}[t]{0.96\columnwidth} \centering
\includegraphics[width=1.0\columnwidth]{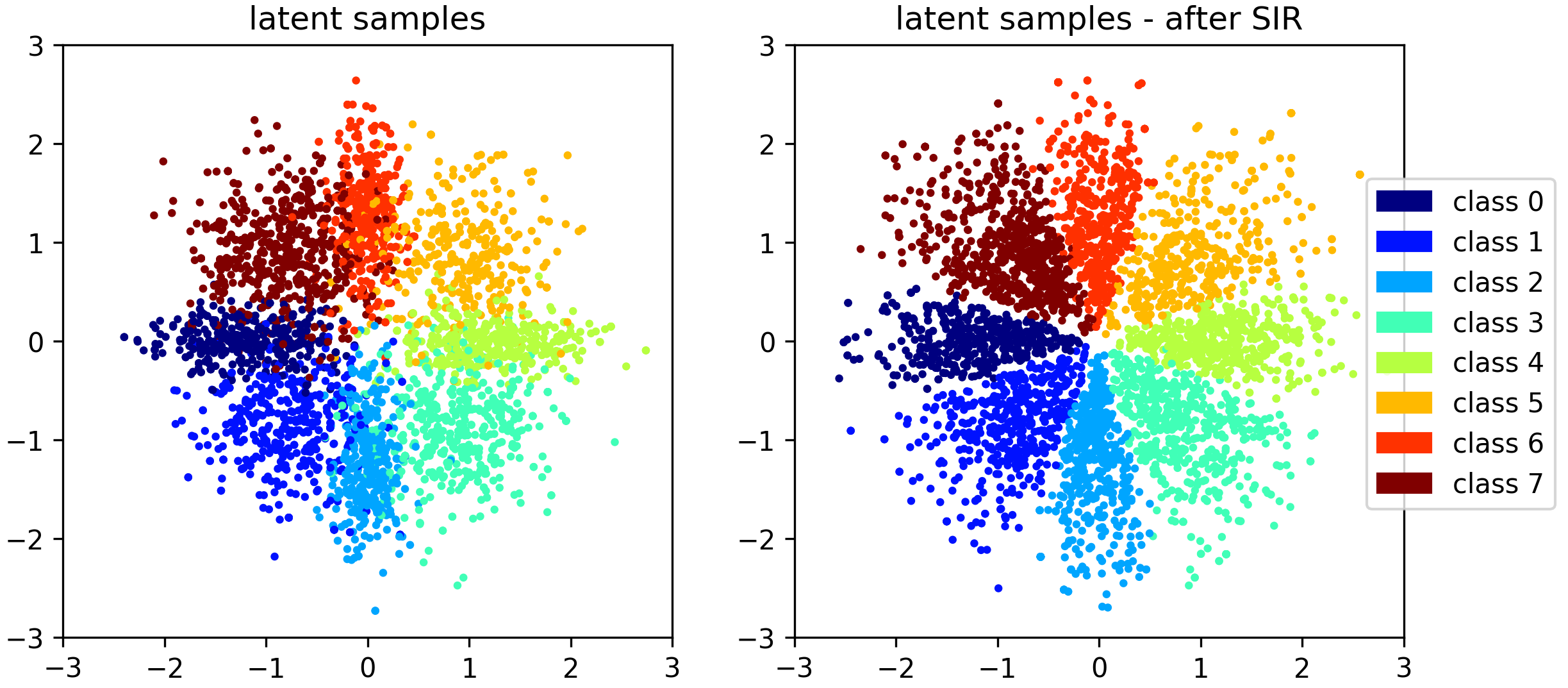}
\caption{Integrated belief and its predictions.
\textbf{Left:} Random samples from integrated variational posterior. Colours correspond to the test points.
\textbf{Right:} Samples after sampling importance re-sampling using all likelihood functions.
}
\label{fig:samples_and_SIR_integrated}
\end{subfigure}
\end{figure}

For the mixture of bi-variate Gaussians dataset, we show latent samples from both information sources in Fig.~\ref{fig:samples_and_SIR_individual}~(left) and samples obtained by sampling importance re-sampling (SIR) using the full likelihood model in \ref{fig:samples_and_SIR_individual}~(right). 
We also show random samples from the integrated beliefs as well as samples obtain by SIR in Fig.~\ref{fig:samples_and_SIR_integrated}~(left) and \ref{fig:samples_and_SIR_integrated}~(right) respectively.
We conclude that the integrated beliefs are much better proposal distributions, resolving the ambiguity of the individual sources. 

\subsection{Visualisation of missing data imputation}
\begin{figure*}\centering
\begin{subfigure}{1.0\textwidth}\centering
	\includegraphics[width=1.0\textwidth]{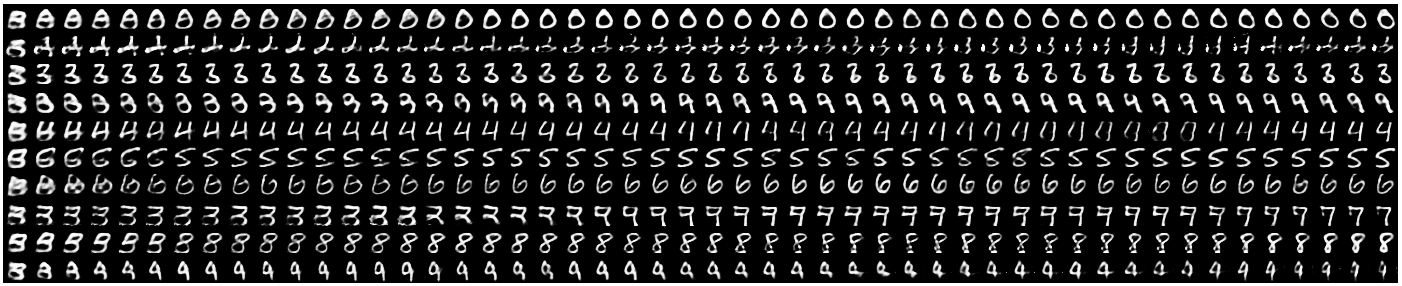}
	\caption{Bottom half of the image is missing}
   \label{fig:missing_data_inference_images_TB}
\end{subfigure}
\qquad
\begin{subfigure}{1.0\textwidth} \centering
	\includegraphics[width=1.0\textwidth]{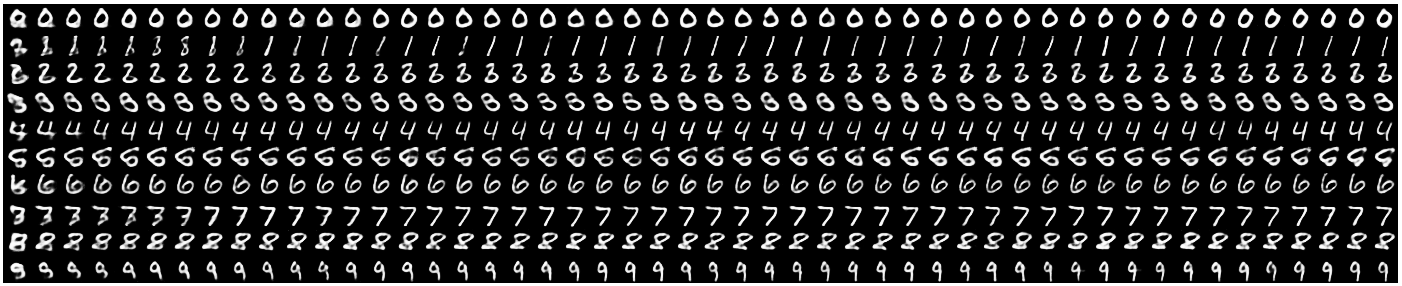}
	\caption{Bottom right quarter of the image is missing}
	\label{fig:missing_data_inference_images_QU}
\end{subfigure}
\caption{Missing data imputation results: Mean of generated images. Observed data (fixed binarised) is kept unchanged, missing data is replaced with randomly generated (binary) image of previous iteration. The initial missing data is drawn randomly from $\operatorname{Ber} \left({0.5}\right)$.
Each of the 10 rows is an exemplar image of digits 0--9.}
\label{fig:missing_data_inference_images}
\end{figure*}

Fig.~\ref{fig:missing_data_inference_images} shows the mean of generated images for 50 steps of the Markov chain procedure for missing data imputation. 
As can be seen in Fig.~\ref{fig:missing_data_inference_images_TB}, the chain does not converge for many digits within 50 steps if too large portions of the data are missing. 
Indeed, we observed that the procedure randomly fails or succeeds to converge for the same input even after 150 steps. 
\subsection{Conditional generations on Caltech-UCSD Birds 200}
\begin{figure*}
\centering
\begin{subfigure}{0.31\textwidth}
	\centering
	\includegraphics[width=1.0\textwidth]{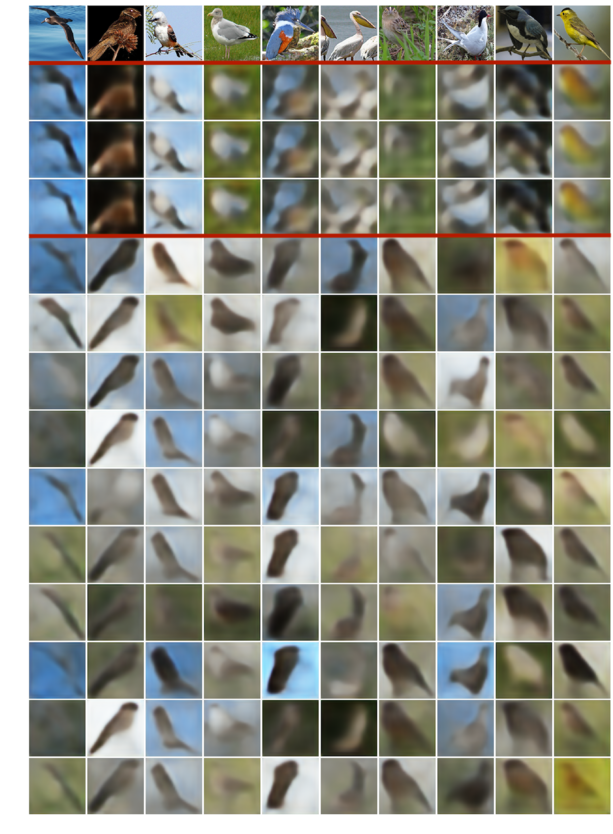}
	\caption{Trained with $\elboa$.}
	\label{fig:generation_cub_msnvi_A}
\end{subfigure}
\begin{subfigure}{0.31\textwidth}
	\centering
	\includegraphics[width=1.0\textwidth]{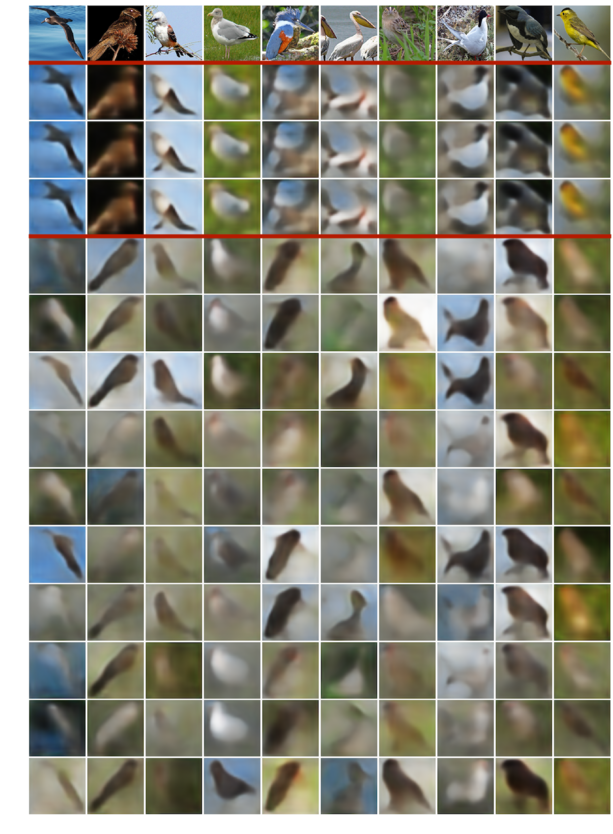}
	\caption{Trained with $\elboac$.}
	\label{fig:generation_cub_msnvi_AC}
\end{subfigure}
\begin{subfigure}{0.31\textwidth}
	\centering
	\includegraphics[width=1.0\textwidth]{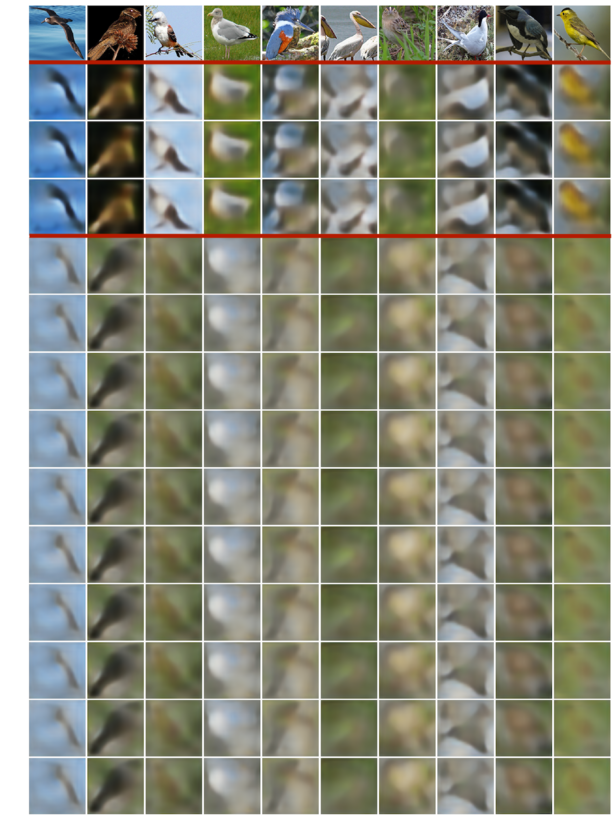}
	\caption{Trained with $\elboac$, K=1.}
	\label{fig:generation_cub_msnvi_AC_K1}
\end{subfigure}
\caption{Conditional image generations, where latent variables are inferred from different sources. \textbf{Row 1:} Target observations. \textbf{Row 2--4:} Latent variables inferred from images. \textbf{Row 5--15:} Latent variables inferred from segmentation masks.}
\label{fig:generation_cub}
\end{figure*}

\begin{figure*}
\centering
\begin{subfigure}{0.31\textwidth}
	\centering
	\includegraphics[width=1.0\textwidth]{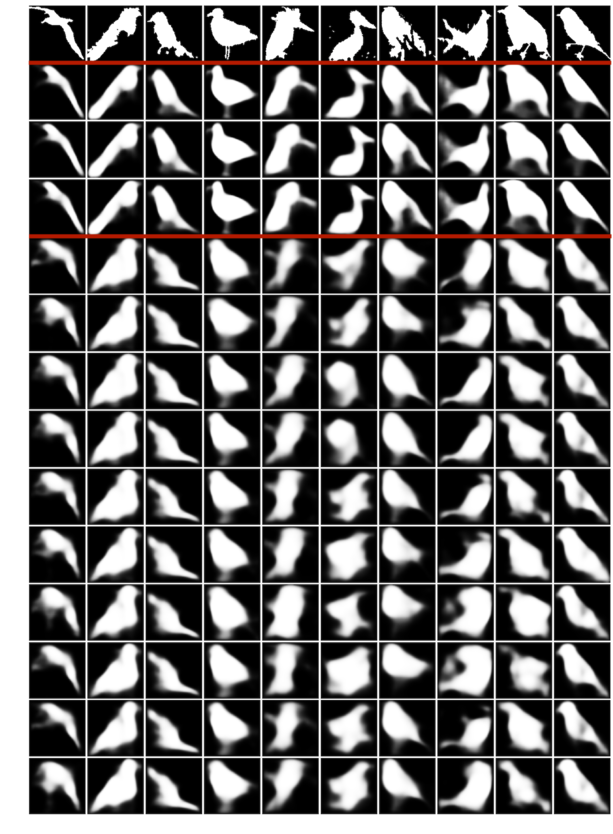}
	\caption{Trained with $\elboa$.}
	\label{fig:generation_cub_segment_msnvi_A}
\end{subfigure}
\begin{subfigure}{0.31\textwidth}
	\centering
	\includegraphics[width=1.0\textwidth]{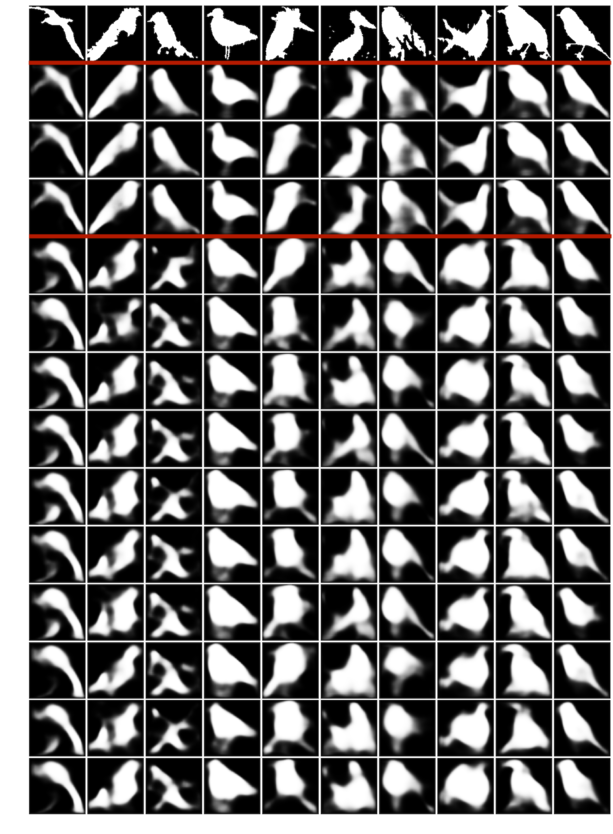}
	\caption{Trained with $\elboac$.}
	\label{fig:generation_cub_segment_msnvi_AC}
\end{subfigure}
\begin{subfigure}{0.31\textwidth}
	\centering
	\includegraphics[width=1.0\textwidth]{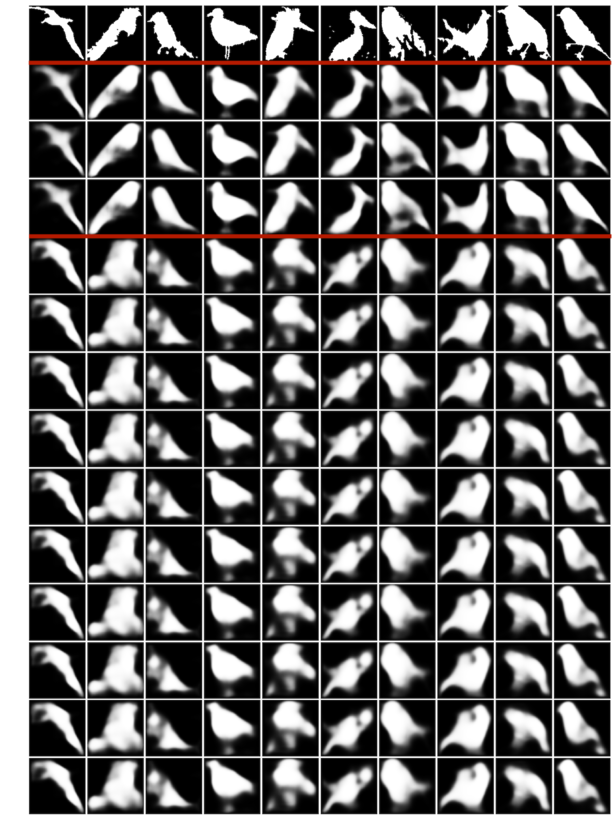}
	\caption{Trained with $\elboac$, K=1.}
	\label{fig:generation_cub_segment_msnvi_AC_K1}
\end{subfigure}
\caption{Conditional segmentation mask generations, where latent variables are inferred from different sources. \textbf{Row 1:} Target observations. \textbf{Row 2--4:} Latent variables inferred from segmentation masks. \textbf{Row 5--15:} Latent variables inferred from images.}
\label{fig:generation_cub_segment}
\end{figure*}

We show conditional generations of images, inferred from images or segmentation masks in Fig.~\ref{fig:generation_cub}. 
When inferring from segmentation masks, the conditional distribution $p_{\theta_\mathrm{img}}(\x_\mathrm{img} \given \x_\mathrm{label})$ should be highly multimodal due to the missing colour information. 
This uncertainty should ideally be covered in the uncertainty of the belief.
As can be seen in Fig.~\ref{fig:generation_cub_msnvi_AC_K1}, learning with a single importance sample leads to predictions of average images.
For completeness, generated segmentation masks are shown in Fig.~\ref{fig:generation_cub_segment}.
\subsection{Experiment setups}\label{sec:appendix:hyperparams}
All inference (generative) models use the same neural network architectures for the different sources, except the first (last) layer, which depends on the dimensions of the data. 
We refer to main parts of the architectures, identical for each source, as ``stem''. 
In case of inference models, the stem is the input to a dense layer with linear (no activation) and sigmoid activations, parameterising the mean and std-dev of the approximate posterior distribution. 
In case of generative models, refer to the respective subsections.

We use the Adam optimiser \cite{corrKingmaB14} with $\beta_{1}=0.9$ and $\beta_{2}=0.999$ in all experiments. 
In the tables, ``dense'' denotes fully connected layers, ``conv'' refers to convolutional layers, ``pool'' refers to pooling (down-sampling), and ``interpol'' refers to a bilinear interpolation (up-sampling). 
$K$ is the number of importance-weighted samples and $D_{z}$ refers to the number of latent dimensions, each modelled with a diagonal normal distribution with zero mean and unit standard deviation. 
\subsubsection{Partially observable mixture of bi-variate Gaussians}
In the pendulum experiment, we use 2 sources, corresponding to the $x$- and $y$-coordinates of the sample from a mixture of bi-variate Gaussians distribution. 
The neural network stems and training hyperparameters are summarised in Tab.~\ref{tab:pogor_architecture}. 
The generative models are both 1D Normal distributions, parameterised by linear dense layers, taking inputs from their respective stems.
\begin{table}\centering
\caption{Neural network architectures (stem) and hyperparameters used for experiments with partially observable mixture of bi-variate Gaussians} 
\label{tab:pogor_architecture}
\begin{minipage}[t]{1.0\columnwidth} \centering
	\begin{tabular}[t]{@{}rrr@{}}
		& \multicolumn{1}{c}{\bf Inference models} & \\
		\toprule
		layer		& activation	& output shape	\\
		\midrule
		dense		&	 tanh		& 32		\\
		dense		&	 tanh		& 32		\\
		\bottomrule
	\end{tabular}
	\end{minipage}
\quad
\begin{minipage}[t]{1.0\columnwidth} \centering
	\begin{tabular}[t]{@{}rrr@{}}
		& \multicolumn{1}{c}{\bf Generative models} & \\
		\toprule
		layer		& activation	& output shape	\\
		\midrule
		dense		&	 tanh		& 32		\\
		dense		&	 tanh		& 32		\\
		\bottomrule
	\end{tabular}
\end{minipage}
\begin{minipage}[t]{1.0\columnwidth} \centering
	\begin{tabular}[t]{@{}rrrrr@{}}
		& \multicolumn{3}{c}{\bf Hyperparameters} & \\
		\toprule
		$K$ & $D_{z}$ & batch size & learning rate & \#iterations \\
		\midrule
		8 & 2 & 32 & 0.0001 & 25k \\
		\bottomrule
	\end{tabular}
\end{minipage}
\end{table}
\subsubsection{MNIST variants}
The neural network stems are summarised in Tab.~\ref{tab:mnist_architecture}. 
The data is modelled as Bernoulli distributions of dimensions 784 for MNIST-NO, 392 for MNIST-TB and 196 for MNIST-QU. 
The Bernoulli parameters are parameterised by linear dense layers, taking inputs from their respective stems.
\begin{table}[h!]\centering
\caption{Neural network architectures and hyperparameters used for experiments with MNIST variants.} 
\label{tab:mnist_architecture}
	\begin{minipage}[t]{1.0\columnwidth} \centering
		\begin{tabular}[t]{@{}rrr@{}}
		& \multicolumn{1}{c}{\bf Inference models} & \\
		\toprule
		layer		& activation	& output shape	\\
		\midrule
		dense		&	 elu		& 200		\\
		dense		&	 elu		& 200		\\
		\bottomrule
		\end{tabular}
	\end{minipage}
\qquad
\begin{minipage}[t]{1.0\columnwidth} \centering
	\begin{tabular}[t]{@{}rrr@{}}
		& \multicolumn{1}{c}{\bf Generative models} & \\
		\toprule
		layer		& activation	& output shape	\\
		\midrule
		dense		&	 elu		& 200		\\
		dense		&	 elu		& 200		\\
		\bottomrule
		\end{tabular}
\end{minipage}
\qquad
\begin{minipage}[t]{1.0\columnwidth} \centering
	\begin{tabular}[t]{@{}rrrrr@{}}
		& \multicolumn{3}{c}{\bf Hyperparameters} & \\
		\toprule
		$K$ & $D_{z}$ & batch size & learning rate & \#iterations \\
		\midrule
		16 & 16 & 128 & 0.00005 & 250k \\
		\bottomrule
		\end{tabular}
\end{minipage}
\end{table}
\subsubsection{Pendulum}
In the pendulum experiment, we use 3 sources with $32\times32$ images for the inference model, but a single observation of $x$- and $y$-coordinates of the pendulum centre. 
The generative model is assumed Normal for both coordinates, where the mean is predicted by a linear dense layer taking inputs from the stem, and the std deviation is a global variable. 
The neural network stems and training hyperparameters are summarised in Tab.~\ref{tab:pendulum_architecture}.
\begin{table}\centering
\caption{Neural network architectures and hyperparameters used for perspective pendulum experiments} 
\label{tab:pendulum_architecture}
\begin{minipage}[t]{1.0\columnwidth} \centering
	\begin{tabular}[t]{@{}rrr@{}}
		& \multicolumn{1}{c}{\bf Inference models} & \\
		\toprule
		layer		& activation	& output shape	\\
		\midrule
		dense		&	 tanh		& 32		\\
		dense		&	 tanh		& 32		\\
		\bottomrule
	\end{tabular}
	\end{minipage}
\quad
\begin{minipage}[t]{1.0\columnwidth} \centering
	\begin{tabular}[t]{@{}rrr@{}}
		& \multicolumn{1}{c}{\bf Generative models} & \\
		\toprule
		layer		& activation	& output shape	\\
		\midrule
		dense		&	 tanh		& 256		\\
		dense		&	 tanh		& 64		\\
		dense		&	 tanh		& 16		\\
		\bottomrule
	\end{tabular}
\end{minipage}
\begin{minipage}[t]{1.0\columnwidth} \centering
	\begin{tabular}[t]{@{}rrrrr@{}}
		& \multicolumn{3}{c}{\bf Hyperparameters} & \\
		\toprule
		$K$ & $D_{z}$ & batch size & learning rate & \#iterations \\
		\midrule
		16 & 2 & 16 & 0.00005 & 50k \\
		\bottomrule
	\end{tabular}
\end{minipage}
\end{table}

\subsubsection{Caltech-UCSD Birds 200}
The neural network stems are summarised in Tab.~\ref{tab:cub_architecture}. 
Images are modelled as diagonal normal distributions and segmentation masks as Bernoulli distributions. 
The generative model stem is the input to a $5\times5$-transposed convolutional layers with stride 2, yielding the mean of the likelihood function. 
The standard deviations are global and shared for all pixels. 
Leaky rectified linear units (lrelu) use $\alpha=0.10$.
\begin{table}[h!]\centering
\caption{Neural network architectures and hyperparameters used for Caltech-UCSD Birds 200 experiments} 
\label{tab:cub_architecture}
\begin{minipage}[t]{1.0\columnwidth} \centering
\begin{tabular}[t]{@{}rrrrr@{}}
& \multicolumn{3}{c}{\bf Inference models} & \\
\toprule
layer & kernel & stride & activation & output shape \\
\midrule
conv  &	$3\!\times\!3$	& 1	&	lrelu	&	128x128x16	\\
conv  &	$3\!\times\!3$	& 1	&	lrelu	&	128x128x16	\\
pool  &	$3\!\times\!3$	& 2	& 	  -		&	64x64x16	\\
conv  &	$3\!\times\!3$	& 1	&	lrelu	&	64x64x32	\\
conv  &	$3\!\times\!3$	& 1	&	lrelu	&	64x64x32	\\
pool  &	$3\!\times\!3$	& 2	& 	  -		&	32x32x32	\\
conv  &	$3\!\times\!3$	& 1	&	lrelu	&	32x32x48	\\
conv  &	$3\!\times\!3$	& 1	&	lrelu	&	32x32x48	\\
pool  &	$3\!\times\!3$	& 2	& 	  -		&	16x16x48	\\
conv  &	$3\!\times\!3$	& 1	&	lrelu	&	16x16x64	\\
conv  &	$3\!\times\!3$	& 1	&	lrelu	&	16x16x64	\\
pool  &	$3\!\times\!3$	& 2	& 	  -		&	8x8x64		\\
conv  &	$3\!\times\!3$	& 1	&	lrelu	&	8x8x96		\\
conv  &	$3\!\times\!3$	& 1	&	lrelu	&	8x8x96		\\
pool  &	$3\!\times\!3$	& 2	& 	  -		&	4x4x96		\\
dense &	 -	& - &	linear	&	256			\\
\bottomrule
\end{tabular}
\end{minipage}
\quad
\begin{minipage}[t]{1.0\columnwidth} \centering
\begin{tabular}[t]{@{}rrrrr@{}}
& \multicolumn{3}{c}{\bf Generative models} & \\
\toprule
layer & kernel & stride & activation & output shape \\
\midrule
dense 	  &	 -	& - &	linear	&	4x4x96		\\
conv  	  &	$3\!\times\!3$	& 1	&	lrelu	&	4x4x64		\\
conv  	  &	$3\!\times\!3$	& 1	&	lrelu	&	4x4x64		\\
interpol  &	$3\!\times\!3$	& 2	& 	  -		&	8x8x64		\\
conv  	  &	$3\!\times\!3$	& 1	&	lrelu	&	8x8x48		\\
conv  	  &	$3\!\times\!3$	& 1	&	lrelu	&	8x8x48		\\
interpol  &	$3\!\times\!3$	& 2	& 	  -		&	16x16x48	\\
conv	 	  &	$3\!\times\!3$	& 1	&	lrelu	&	16x16x32	\\
conv	  	  &	$3\!\times\!3$	& 1	&	lrelu	&	16x16x32	\\
interpol  &	$3\!\times\!3$	& 2	& 	  -		&	32x32x32	\\
conv		  &	$3\!\times\!3$	& 1	&	lrelu	&	32x32x16	\\
conv	  	  &	$3\!\times\!3$	& 1	&	lrelu	&	32x32x16	\\
interpol  &	$3\!\times\!3$	& 2	& 	  -		&	64x64x16	\\
\bottomrule
\end{tabular}
\end{minipage}
\begin{minipage}[t]{1.0\columnwidth} \centering
\begin{tabular}[t]{@{}rrrrr@{}}
& \multicolumn{3}{c}{\bf Hyperparameters} & \\
\toprule
$K$ & $D_{z}$ & batch size & learning rate & \#iterations \\
\midrule
80 & 96 & 16 & 0.0002 & 25k \\
\bottomrule
\end{tabular}
\end{minipage}
\end{table}
\end{document}